\documentclass[10pt,twocolumn,letterpaper]{article}

\usepackage[pagenumbers]{cvpr}            
\usepackage{amsbsy}
\usepackage{amsmath}
\usepackage{multirow}
\usepackage{float}
\usepackage[para]{footmisc}
\usepackage[accsupp]{axessibility}

\usepackage{amsmath}
\usepackage{amssymb}
\usepackage{comment}

\definecolor{frenchblue}{rgb}{0.0, 0.45, 0.73}
\definecolor{gold}{rgb}{0.83, 0.69, 0.22}

\newcommand{\datasetname}{4D-DRESS}

\definecolor{cvprblue}{rgb}{0.21,0.49,0.74}
\usepackage[pagebackref,breaklinks,colorlinks,citecolor=cvprblue]{hyperref}

\def\paperID{***} 
\def\confName{CVPR}
\def\confYear{2024}

\title{4D-DRESS: A 4D Dataset of \\
Real-World Human Clothing With Semantic Annotations}

\author{Wenbo Wang$^{*1}$~~
\and
Hsuan-I Ho$^{*1}$~~
\and
Chen Guo$^{1}$~~
\and
Boxiang Rong$^{1}$~~
\and
Artur Grigorev$^{1,2}$~~
\and
Jie Song$^{1}$~
\and
Juan Jose Zarate$^{\dagger1}$~
\and
Otmar Hilliges$^{1}$~
\vspace{0.1em}
\and
~~~~~~~~Department of Computer Science, ETH Zürich~~~~~~~~
\and
~~~~~~~~Max Planck Institute for Intelligent Systems, Tübingen~~~~~~~~
\\
{\small\url{https://ait.ethz.ch/4d-dress}}
}
\begin{document}
\twocolumn[{%
\renewcommand\twocolumn[1][]{#1}%
\maketitle
\begin{center}
    \centering
    \captionsetup{type=figure}
    \includegraphics[width=0.95\linewidth]{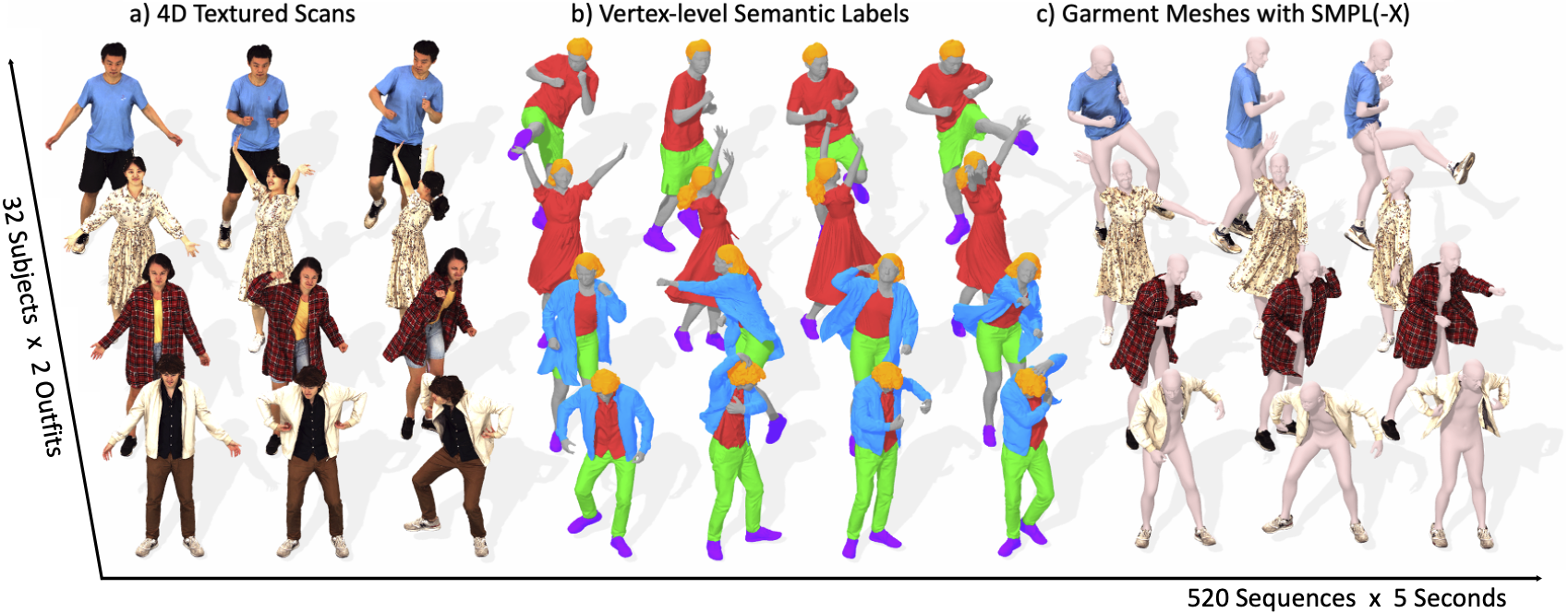}
    \vspace{-0.8em} 
    \caption{\textbf{Overview of \datasetname.} We propose the first real-world 4D dataset of human clothing, capturing 64 human outfits in more than 520 motion sequences. These sequences include a) high-quality 4D textured scans; for each scan, we annotate b) vertex-level semantic labels, thereby obtaining c) the corresponding garment meshes and fitted SMPL(-X) body meshes.
    }
    \label{fig:teaser}
\end{center}%
}]
\def\thefootnote{*}\footnotetext{Equal contributors}\def\thefootnote{$\dagger$}\footnotetext{Corresponding author}
\def\thefootnote{\arabic{footnote}}

\begin{abstract}
The studies of human clothing for digital avatars have predominantly relied on synthetic datasets.
While easy to collect, synthetic data often fall short in realism and fail to capture authentic clothing dynamics.
Addressing this gap, we introduce \datasetname, the first real-world 4D dataset advancing human clothing research with its high-quality 4D textured scans and garment meshes.
\datasetname~captures 64 outfits in 520 human motion sequences, amounting to 78k textured scans.
Creating a real-world clothing dataset is challenging, particularly in annotating and segmenting the extensive and complex 4D human scans.
To address this, we develop a semi-automatic 4D human parsing pipeline.
We efficiently combine a human-in-the-loop process with automation to accurately label 4D scans in diverse garments and body movements.
Leveraging precise annotations and high-quality garment meshes, we establish several benchmarks for clothing simulation and reconstruction. 
\datasetname~offers realistic and challenging data that complements synthetic sources, paving the way for advancements in research of lifelike human clothing.

\end{abstract}


\definecolor{Gray}{gray}{0.85}
\renewcommand{\arraystretch}{1.05}
\begin{table*}[t]
    \centering
    \footnotesize
\begin{tabular}{l|c|c|c|c|c|c}
\toprule
    Dataset & \# of  Outfits & \# of Frames & Data Format & Textured &   \begin{tabular}[c]{@{}c@{}} Semantic \\ Labels \end{tabular} &
    \begin{tabular}[c]{@{}c@{}} Loose \\ Garments \end{tabular}  \\
\midrule
    \rowcolor{Gray} TailorNet~\cite{TailorNet} & 9 & 5.5k & SMPL + Garments &  & \checkmark & \\
    \rowcolor{Gray} ReSynth~\cite{POP} & 24 & 30k & SMPLX + Point Clouds  &  &  & \checkmark  \\
    \rowcolor{Gray} CLOTH3D~\cite{CLOTH3D} & 8.5k & 2.1M & SMPL + Garments & \checkmark & \checkmark & \checkmark  \\
    \rowcolor{Gray} CLOTH4D~\cite{CLOTH4D} & 1k & 100k & Mesh + Garments & \checkmark & \checkmark & \checkmark \\
    \rowcolor{Gray} BEDLAM~\cite{BEDLAM} & 111 & 380k & SMPL-X + Garments & \checkmark & \checkmark & \checkmark\\
    \rowcolor{Gray} D-LAYERS~\cite{D-LAYERS} & 5k & 700k & SMPL + Garments &  & \checkmark & \checkmark \\
\midrule
    BUFF~\cite{BUFF} & 6 & 14k & Scans + SMPL &  \checkmark &   &   \\
    CAPE~\cite{CAPE} & 15 & 140k & SMPL+D &  &   &   \\
    ActorsHQ~\cite{HumanRF} & 8 & 39k & Scans &  &  & \checkmark \\
    X-Humans~\cite{X-Avatar} & 20 & 35k & Scans + SMPL-(X) & \checkmark &  & \\
    4DHumanOutfit~\cite{4DHumanOutfit} & 14 & 459k & Scans + SMPL & \checkmark &  & \checkmark \\
\midrule
    \bf \datasetname~(Ours) &  \bf 64  & \bf 78k &  \bf
    \begin{tabular}[c]{@{}c@{}} \textbf{Scans + SMPL(-X)} \\ + \textbf{Garments} \end{tabular} &  \bf \checkmark  & \bf \checkmark  & \bf \checkmark  \\
\bottomrule
\end{tabular}
\vspace{-0.5em}
\caption{\textbf{Summary of 4D clothed human datasets}. The datasets highlighted in \colorbox{Gray}{gray color} are synthetic datasets while the others are real-world scans. \# of Outfits: number of outfits included; \# of Frames: total number of 3D human frames; Data Format: 3D representations of human bodies and garments; Textured: with textured map or not; Semantic Labels: with semantic labels for clothing or not; Loose Garments: containing challenging loose clothing such as dresses or not. \datasetname~demonstrates outstanding features against others.
}
  \vspace{-1.5em}
\label{tab:dataset}
\end{table*}

\vspace{-1.5em}
\section{Introduction}

Human clothing is crucial in various applications such as 3D games, animations, and virtual try-on. 
Researchers are actively investigating algorithms for clothing reconstruction~\cite{jiang2020bcnet,corona2021smplicit,Moon_2022_ECCV_ClothWild} and simulation~\cite{HOOD, PBNS, NCS}, to achieve realistic clothing behavior, enhance user engagement, and enable cross-industry applications.
These algorithms are frequently developed and assessed using synthetic datasets~\cite{CLOTH3D,CLOTH4D,BEDLAM}, since they comprise a) meshes covering various garment types and outfits and b) parametric body models with diverse motions.
While synthetic datasets lead in outfit quantity and the number of frames provided (refer to~\cref{tab:dataset}), there also presents a significant challenge in bridging the domain gap between the synthetic and real garments.
Despite the recently released real-world 4D human datasets such as X-Humans~\cite{X-Avatar}, ActorsHQ~\cite{HumanRF}, and 4DHumanOutfit~\cite{4DHumanOutfit}, a key limitation persists: they lack accurately segmented garment meshes, offering only raw human scans.
Moreover, these datasets are limited in the number of loose garments (e.g., jackets and dresses) or dynamic motions, which reduces their applicability as test benches.
These challenges highlight the need for a real-world 4D dataset that provides semantic annotations and captures diverse garments across various body motions.

In this work, we contribute \datasetname, the first real-world dataset of human clothing with 4D semantic segmentation.
We aim to provide an evaluation testbench with real-world data for tasks related to human clothing in computer vision and graphics.
We capture over 520 human motion sequences featuring 64 distinct real-world human outfits in a high-end multi-view volumetric capture system, similar to the one used in~\cite{collet2015high}.
The complete dataset comprises a total of 78k frames, each composed of an 80k-face triangle mesh, a 1k resolution textured map, and a set of 1k resolution multi-view images.
As illustrated in~\cref{fig:teaser}, we provide a) high-quality 4D textured scans, b) vertex-level semantic labels for various clothing types, such as upper, lower, and outer garments, and c) garment meshes along with their registered SMPL(-X) body models.

Capturing real-world 4D sequences of humans wearing various clothing and performing diverse motions requires dedicated high-end capture facilities.
Moreover, processing these clips into accurately annotated and segmented 4D human scans presents significant challenges. 
To develop our dataset, we tackled the task of labeling 78k high-resolution meshes at the vertex level. 
Given that the mesh topologies of consecutive frames do not inherently correspond, consistently propagating 3D vertex labels from one frame to the next is non-trivial. 
While previous methods~\cite{ClothCap,MGN} attempted to fit a fixed-topology parametric body model to the scans, these template-based approaches still struggle with scenarios such as a jacket being lifted to reveal a shirt or the emergence of new vertices on a flowing coat as illustrated in the example shown in~\cref{fig:ablation}.
Consequently, we opted for an alternative approach. 
We developed a semi-automatic and template-free 4D human parsing pipeline. 
Leveraging semantic maps from a 2D human parser~\cite{Graphonomy} and a segmentation model~\cite{SAM}, we extended these techniques to 4D, considering both multi-view and temporal consistency. 
Our pipeline accurately assigns vertex labels without manual intervention in 96.8\% of frames. 
Within the remaining scans, only 1.5\% of vertices require further rectification, addressed via a human-in-the-loop process.

The quality of the ground-truth data in \datasetname~allows us to establish several evaluation benchmarks for diverse tasks, including clothing simulation, reconstruction, and human parsing. Our evaluation and analysis demonstrate that \datasetname~offers realistic and challenging human clothing that cannot be readily modeled by existing algorithms, thereby opening avenues for further research. In summary, our contributions include:
\begin{itemize}
    \item the first real-world 4D human clothing dataset comprising 4D textured scans, vertex-level semantic labels, garment meshes, and corresponding parametric body meshes.
    \item a semi-automatic and template-free 4D human parsing pipeline for efficient data annotation.
    \item evaluation benchmarks showing the utility of our dataset.
\end{itemize}

\section{Related Work}

\paragraph{4D clothed human dataset.} 
\vspace{-.5em}
Datasets featuring clothed humans can be divided into two categories.
Firstly, synthetic datasets~\cite{POP,TailorNet,CLOTH3D,CLOTH4D,BEDLAM,D-LAYERS} create large volume of synthetic data using graphic engines~\cite{unrealengin5} and simulation tools~\cite{CLO} (\cref{tab:dataset} top).
These datasets are easy to scale with ground truth semantic labels available by design.
However, they often lack realism in human appearances, clothing deformations, and motion dynamics.
Even though recent work~\cite{BEDLAM,wood2021fake} attempted to achieve photorealistic human textures with manual efforts, it is challenging to precisely mimic the way real-world clothing moves and deforms.
Therefore, it is essential to create datasets of real-world human clothing by capturing these intricate details.

The second category (\cref{tab:dataset} bottom) involves using multi-view volumetric capture systems~\cite{Joo_2017_TPAMI,collet2015high} to collect datasets of people dressed in real-world clothing~\cite{CAPE,BUFF,X-Avatar,4DHumanOutfit,EditableAvatar,2K2K,SIZER,DeepFashion3D,tao2021function4d, HumanRF,DeepCloth}.
However, the resources required for capturing, storing, and processing this data are substantial, which limits the size of these publicly available datasets~\cite{BUFF,X-Avatar,4DHumanOutfit}. 
Moreover, these methods do not inherently provide labeled annotations, offering only temporally uncorrelated scans.
This makes the raw data on these datasets less suitable for research focusing on human clothing.
\datasetname~gathers a variety of human subjects and outfits providing accurate semantic labels of human clothing, garment meshes, and SMPL/SMPL-X fits.

\vspace{-1.5em}
\paragraph{Human parsing.}
Human parsing~\cite{yang2023humanparsing} is a specific task within semantic segmentation aimed at identifying detailed body parts and clothing labels. 
Conventionally, this challenge is tackled using deep neural networks, trained on images with their corresponding semantic labels~\cite{LIP,CIHP,PASCAL}.
Although these methods have been successful in 2D~\cite{Graphonomy,SCHP,CDGNet,Wang_2020_CVPR,he2020grapy,he2021progressive}, applying them to annotate 3D and 4D scans is still a challenge. 
Previous work has explored it using two distinct strategies. 
One strategy, used by SIZER~\cite{SIZER} and MGN~\cite{MGN}, involves rendering multi-view images and projecting parsing labels onto 3D meshes through a voting process. 
While this method considers consistency across multiple views, it overlooks temporal consistency and falls short of accurately labeling 4D scans. 
Another approach, used by ClothCap~\cite{ClothCap}, registers all scans to a fixed-topology SMPL model~\cite{SMPL} with per-vertex displacements. 
Yet, this method struggles with handling large motions and complex clothing due to limited template resolutions and model-fitting capabilities. 
This results in noisy labels near boundaries and loose garments.
In contrast, our approach combines multi-view voting and optical warping in a template-free pipeline, achieving both multi-view and temporal consistency.
\begin{figure*}[t]
  \centering
  \includegraphics[width=0.88\linewidth]{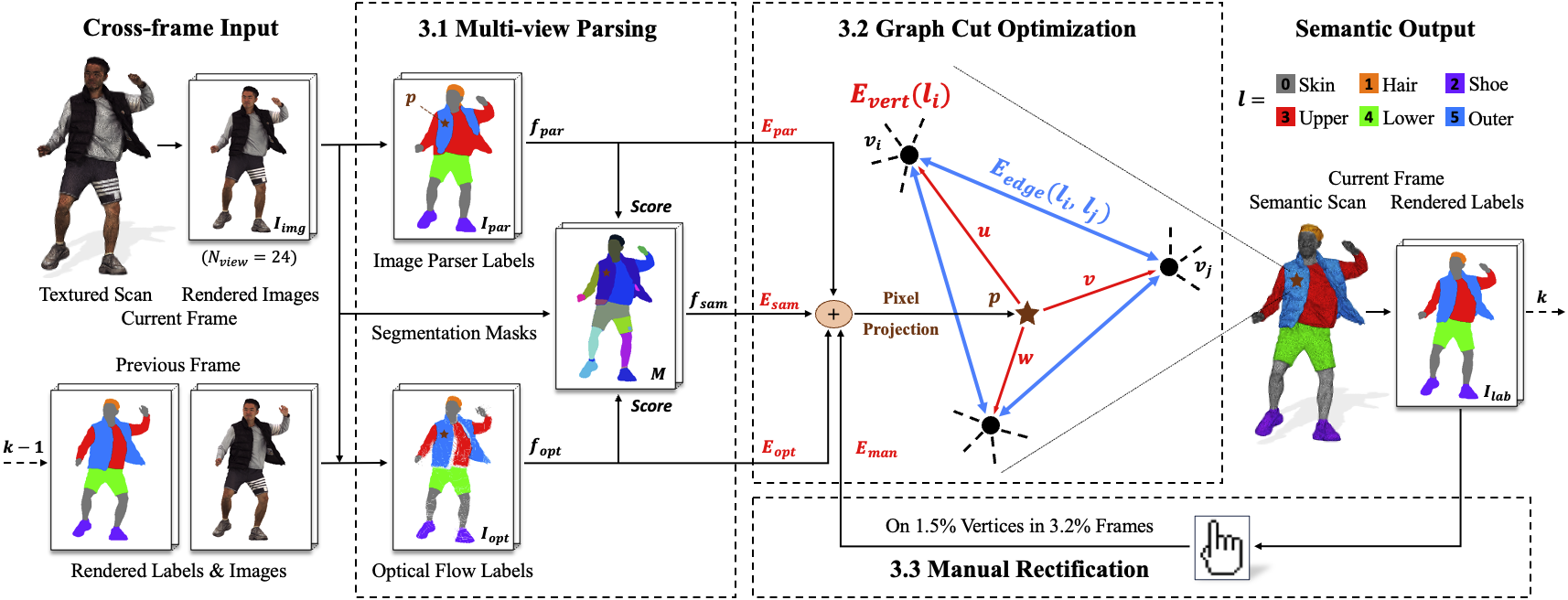}
  \vspace{-0.5em}
  \caption{{\bf 4D Human parsing method.} We first render current and previous frame scans into multi-view images and labels. Then collect multi-view parsing results from the image parser, optical flows, and segmentation masks (\cref{sec:multi-view_parsing}). Finally, we project multi-view labels to 3D vertices and optimize vertex labels using the Graph Cut algorithm with vertex-wise unary energy and edge-wise binary energy (\cref{sec:graph_cut}). The manual rectification labels can be easily introduced by checking multi-view rendered labels. (\cref{sec:manual_efforts}).}
  \label{fig:method}
  \vspace{-1.em}
\end{figure*}

\vspace{-0.5em}
\section{Methodology}

\vspace{-0.5em}
To accurately label each vertex within our 4D textured scan sequences, we leverage a semi-automatic parsing pipeline that incorporates but minimizes manual efforts during the labeling process. \cref{fig:method} depicts the overall workflow of our pipeline.
We first render 24 multi-view images of the current frame textured scan.
We combine those images with the previous frame's multi-view images and labels to deploy three state-of-the-art tools to vote candidate labels for each rendered pixel (\cref{sec:multi-view_parsing}): a) human image parser, b) optical flow transfer, and c) segmentation masks. 
Next, we re-project and fuse all the 2D label votes via a Graph Cut optimization to obtain vertex-level semantic labels, considering neighboring and temporal consistency (\cref{sec:graph_cut}).
For those challenging frames where further labeling refinement is needed (around 3\% in our dataset), we refined their semantic labels with a manual rectification step that we feed back into the optimization (\cref{sec:manual_efforts}).
We describe the details of the pipeline within this section.

\subsection{Multi-view Parsing}
\vspace{-0.5em}
\label{sec:multi-view_parsing}
At each frame $k \in \left\{ 1,...,{N}_{frame}\right\}$, we render the 3D-mesh into a set of multi-view images, consisting of twelve horizontal, six upper, and six lower uniformly distributed views. We note this as $I_{img, n, k}$ with $n \in \left\{ 1,...,{N}_{view}=24\right\}$.
Within the multi-view space, we tackle the problem of assigning a label vote $l$ to each pixel $p$ using multi-view image-based models.
The label $l$ varies for human skin, hair, shoes, upper clothing (shirts, hoodies), lower clothing (shorts, pants), and outer clothing (jackets, coats). 
For clarity, we omit the frame index ($k$) in the following unless they are strictly needed. 
Please refer to \cref{fig:method} and the Supp. Mat. for more label definitions and the versatility of our parsing method with new labels like belts and socks.

\vspace{-1.1em}
\paragraph{Human image parser (\emph{PAR}).}
Our primary source of labels is a deep-learning image parser, which provides pixel-level votes for body parts and clothes. Specifically, we apply Graphonomy \cite{Graphonomy} to each view $n$ and store the labels as a new set of images $\{I_{par}\}$ (see \cref{fig:method}).
These labels are then accessible by the vote function $f_{par, n}(p, l)$ that checks if the image $I_{par,n}$ matches the value $l$ at the pixel $p$, in which case returns 1, or 0 otherwise. 
This vote function and the other two defined below will be crucial later when setting our full-mesh optimization (\cref{sec:graph_cut}).

\vspace{-1.1em}
\paragraph{Optical flow transfer (\emph{OPT}).}
This block leverages the previous frame's multi-view labels to provide temporal consistency. 
Specifically, we use the optical flow predictor RAFT \cite{RAFT} to transfer multi-view labels in the $k-1$ frame to the current $k$ frame using the texture features on the rendered multi-view images.
Similarly to the image parser above, the optical flow output goes to a set $\{I_{opt}\}$. 
These labels are accessible via the vote function $f_{opt, n}(p,l)$, which checks $I_{opt,n}$ and returns 1 if label $l$ is in $p$ and 0 otherwise.

\begin{figure*}[th]
\centering
\includegraphics[width=0.9\linewidth]{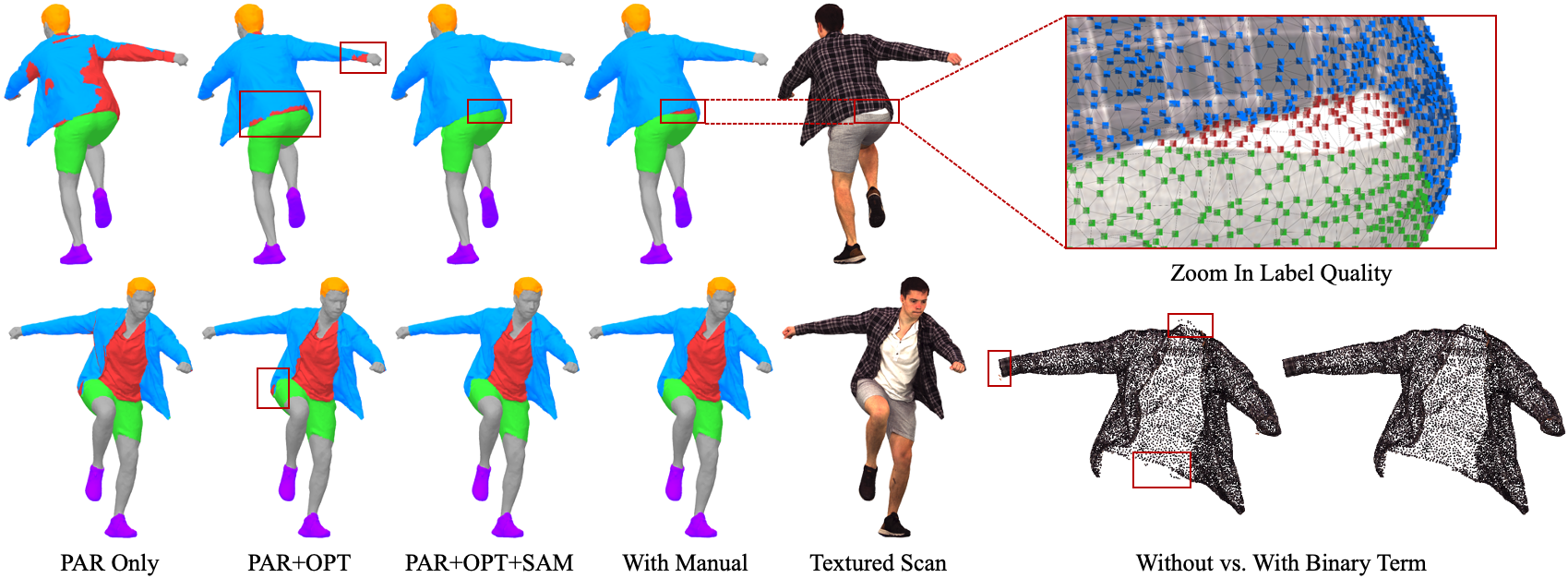}
\vspace{-0.6em}
\caption{\textbf{Qualitative ablation study.} We visualize the effectiveness of our 4D human parsing method on our {\datasetname} dataset. From left to right, we show the improvements after adding the optical flow labels and mask scores to the multi-view image parser labels. The manual rectification efforts can be easily introduced from multi-view rendered labels, with which we achieve high-quality vertex annotations. The problem of isolated labels can be relieved by introducing the edge-wise binary energy term.}
\label{fig:ablation}
  \vspace{-1.em}
\end{figure*}

\vspace{-1.1em}
\paragraph{Segmentation masks (\emph{SAM}).}
The multi-view votes generated by the Human Image Parser sometimes lack 3D consistency, particularly when dealing with open garments under dynamic motions (cf. \cref{fig:ablation}). 
While the votes derived from the optical flows provide a cross-frame prior, they may not accurately track every human part and can't identify newly emerging regions. 
Therefore, we introduce segmentation masks to regularize the label consistency within each masked region. 
We apply the \emph{Segment Anything Model} \cite{SAM} to each rendered image and obtain a self-define group of masks $ M_{m,n}$, with the index $m \in \left\{ 1, ...N_{mask,n} \right\}$.
Within a mask $M_{m,n}$ we compute the score function $\mathcal{S} (l, M_{m, n})$ that fuses the votes of the image parser and the optical flow, normalized by the area of the mask:

\begin{equation}
    \mathcal{S} (l, M_{m, n}) = 
    \frac{
    \underset{p \in M_{m,n}}{\sum} \left[ f_{par,n}(p,l) + \lambda_{po} f_{opt,n}(p,l) \right] 
    }{
    \underset{p \in M_{m,n}}{\sum} \left( 1 + \lambda_{po} \right)
    },
\end{equation}

\noindent where the factor $\lambda_{po}$ weights the contribution of \emph{OPT} over \emph{PAR}.
We now define a check function, $\mathcal{C}(p, M_{m, n})$, that returns 1 if the input evaluation pixel $p$ is in the mask $M_{m, n}$ and 0 otherwise.
Finally, we obtain the corresponding vote function by summing over all the masks in the image:
\begin{equation} \label{eq:f_sam}
    f_{sam, n}(p,l) = \underset{m \in 1:N_{mask,n}}{\sum} \mathcal{C}(p, M_{m, n}) * \mathcal{S}(l, M_{m,n}).
\end{equation}

\vspace{-0.5em}
\subsection{Graph Cut Optimization for Vertex Parsing}
\label{sec:graph_cut}
\vspace{-.5em}
The next step in our semi-automatic process is combining all the labels obtained in \cref{sec:multi-view_parsing} to assign a unique label to each scan vertex ${v}_{i}$, with $i \in \left\{ 1,...,{N}_{vert} \right\}$. 
We frame this 3D semantic segmentation problem as a graph cut optimization: each 3D frame is interpreted as a graph $G$, where vertices are now nodes and mesh edges are connections.
Note that in a traditional Graph Cut, the values of the nodes are fixed, and the optimization computes only the cost of breaking a connection. In our case, we have several \emph{votes} for a vertex label, coming from three different tools and from concurrent multi-view projections.
We define our cost function that consists of two terms,
\begin{equation}
    E(L)= \underset{i \in 1:N_{vert}}{\sum} E_{vert}(l_{i}) + \underset{i,j \in 1:N_{vert}}{\sum} E_{edge}(l_{i}, l_{j}),
\end{equation}
\noindent where $ L = \left\{ {l}_{i} \right\} $ represents all the vertex labels in current frame. As described below, the term $E_{vert}$ combines the different votes into a single cost function, while $E_{edge}$ evaluates neighboring labels for consistent 3D segmentation. 
We follow an approach similar to \cite{GraphCut}.

\vspace{-1.5em}
\paragraph{Vertex-wise unary energy.} 
The cost function per node or \emph{Unary} energy comes from combining the different votes obtained in the multi-view image processing (see \cref{sec:multi-view_parsing}):
\begin{equation}\label{eq:E_node}
    E_{vert}(l_{i})=  \underset{n \in 1:N_{view}}{\sum} \frac{\lambda_{p}E_{par, n} + \lambda_{o}E_{opt, n} + \lambda_{s}E_{sam, n}}{N_{view}},
\end{equation}

\noindent where we combine the human image parser ($ E_{par} $), the cross-frame optical prior ($ E_{opt} $), and the segmentation masks regularization ($ E_{sam} $) contributions.
All these energy terms can be written with the same equation by using the notation $\mathcal{X} = \{par, \ opt, \ sam\}$:
\begin{equation}
    {E_{\mathcal{X}, n}}(l_i) = \underset{p \in P(v_i,n)}{\sum} - w_{\mathcal{X}}(p,v_i) \ f_{\mathcal{X},n}(p,l_i) \label{eq:E_generic},
\end{equation}
meaning that energy of the method $\mathcal{X}$, calculated for a proposed label $l_i$, is obtained by summing over those pixels $p \in P(v_i,n)$ whose projections are within a triangle of $v_i$. 
The weights for the cases of $E_{par}$ and $E_{opt}$ are set to the barycentric distance from the projected pixel $p$ to the vertex $v_i$,
which means $w_{par} = w_{opt} = u$ as in \cref{fig:method}. 
For $E_{sam}$ instead, we set the weight $w_{sam}$ to the constant value 1 given that we look for an across-vertex regularization.
\vspace{-1.em}
\paragraph{Edge-wise binary energy.}
The \emph{Binary} energy term penalizes the case of adjacent vertices with different labels, encouraging neighboring vertices to take the same label. 
Being $A$ the adjacency matrix of the graph $G$ and $\delta$ the Dirac delta function, the edge cost can be calculated as follows:
\begin{equation}
    E_{edge}(l_{i}, l_{j})= \lambda_{b} \ A_{i, j} \ (1 - \delta(l_{i}, l_{j})),
\end{equation}
which increases the energy by $\lambda_{b}$ in the case that the adjacent vertices $ v_{i}, v_{j} $ take different labels $ l_{i} \neq l_{j} $.

\subsection{Manual Rectification of 3D Labels}
\label{sec:manual_efforts}
\vspace{-.5em}
When manual rectification is needed, we introduce it back into the multi-view space as an additional 2D annotation, and we recalculate the steps in \cref{sec:graph_cut}. 
Concretely, we ran the graph cut optimization for the first time.
Then, we rendered the vertex labels into multi-view labels, from which we let a person introduce corrections by comparing the resulting labels with the textured multi-view images.
Similarly to the vote functions of the image parser and optical flow, we create a vote function $f_{man}(p,l)$ that accesses this set of images with rectified annotations and returns 1 if the label $l$ is assigned to the pixel $p$ and 0 otherwise. 

Similar to previous cases, we define a per-view manual energy ($E_{man}$) by using the variable $\mathcal{X} = man$ in \cref{eq:E_generic}, and we added it to the global per-node energy $E_{vert}$ in \cref{eq:E_node}.
We use a constant large weight for $w_{man}$ to favor the manual annotation over other sources of voting where we rectified the labels. 
The final vertex labels $ L^* = \left\{ {l^*}_{i} \right\} $ are obtained after the second round of graph cut optimization. This manual rectification process finally changed 1.5\% of vertices within 3.2\% of all frames. The rectification process is detailed in Supp. Mat.

\begin{table}[t]
    \centering
    \small
\begin{tabular}{l|c|c|c}
\toprule
     &  \multicolumn{1}{c|}{CLOTH4D~\cite{CLOTH4D}} & \multicolumn{2}{c}{BEDLAM}~\cite{BEDLAM} \\
 \midrule
 Method   &  Inner & Inner & Outer \\
 \midrule
 SMPL+D \cite{ClothCap} & 0.872 & 0.846 &   0.765 \\ 
\midrule
 PAR Only \cite{SIZER} & 0.961 & 0.910 & 0.714 \\
 PAR+OPT & 0.969 & 0.963 & 0.942 \\
 PAR+OPT+SAM & {\bf 0.995} & {\bf 0.993} & {\bf 0.988} \\
 
\bottomrule
\end{tabular}
\vspace{-0.8em}
\caption{\textbf{Baseline and ablation study}. Mean accuracy of 4D human parsing methods applied on synthetic datasets. The {\bf Inner} and {\bf Outer} outfits are selected according to our definition in~\cref{sec:dataset}
}
  \vspace{-1.5em}

\label{tab:ablation}
\end{table}

\begin{figure*}[t]
\centering

\includegraphics[width=0.95\linewidth]{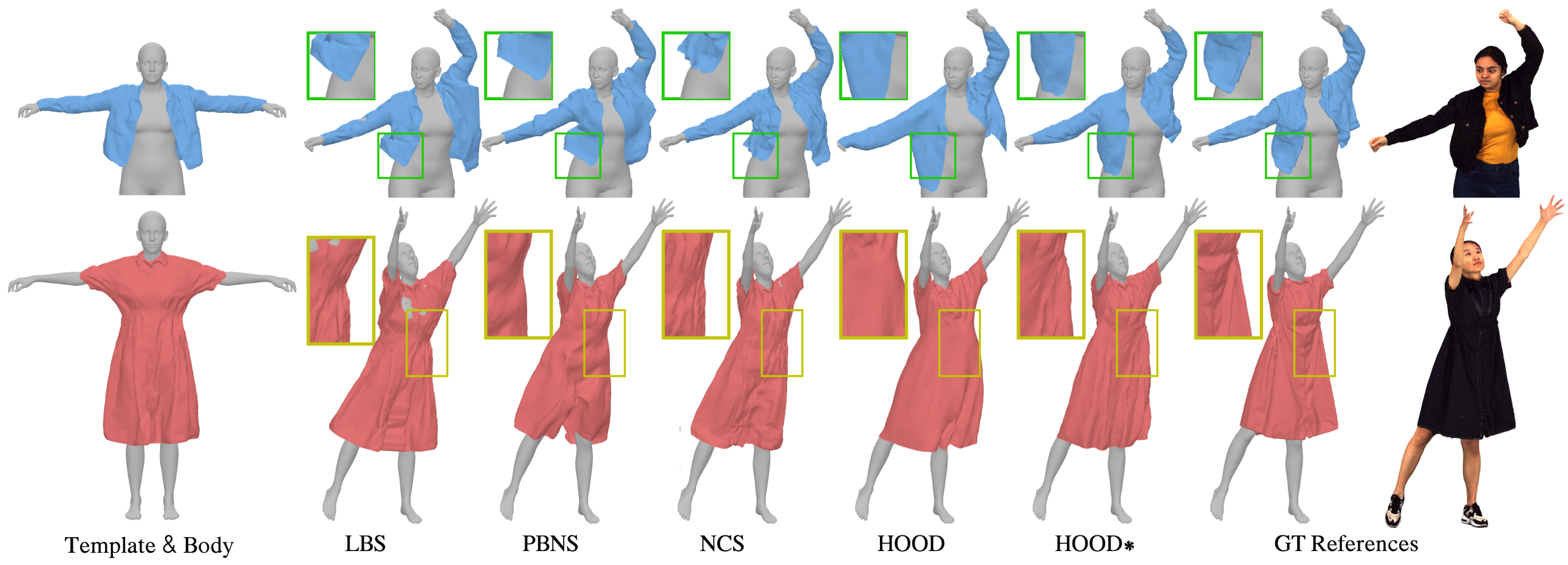}
  \vspace{-1.em}
\caption{\textbf{Qualitative examples for clothing simulation methods}. On the left are templates used for simulations. On the right are ground-truth geometries and original scans, LBS baseline results in body penetrations and overly stretched areas. Compared to other methods, HOOD better models dresses and jackets and, with tuned material parameters, HOOD* achieves simulations closest to the ground truth.}
\label{fig:simulation}
  \vspace{-1.5em}
\end{figure*}

\vspace{-0.5em}
\section{Experiments}
\label{sec:experiments}
\vspace{-.5em}
To validate the effectiveness of our method, we conducted controlled experiments on two synthetic datasets, \textbf{CLOTH4D}~\cite{CLOTH4D} and \textbf{BEDLAM}~\cite{BEDLAM}, where ground-truth semantic labels are available. 
We first compare our parsing method with a template-based baseline~\cite{ClothCap}, that uses a semantic template (SMPL model with per-vertex displacements) to track and parse the clothed human scans.
Due to the limited resolution and the fixed topology nature of the SMPL+D model, its parsing accuracy is lower than 90\% on all synthetic outfits (see~\cref{tab:ablation}). 

We then compare our 4D parsing pipeline with several ablations and report them in~\cref{tab:ablation}. 
We use an example scan from {\datasetname} to support the visualization of the ablation study in~\cref{fig:ablation}.
Using PAR only shows reasonable results for upper and lower clothes. 
Yet, it predicts inconsistent labels at open garments like jackets and coats (\cref{fig:ablation} PAR Only), resulting in only 71.4\% parsing accuracy on the BEDLAM dataset.
The optical flow labels from the previous frame can serve as a cross-frame prior, yet accuracy may vary, particularly in fast-moving arms and cloth boundaries (\cref{fig:ablation} PAR+OPT). 
By fusing both of the previous multi-view labels via the segmentation masks, we achieve better boundary labels (\cref{fig:ablation} PAR+OPT+SAM), with 98.8\% accuracy on the outer outfits in BEDLAM, with challenging open garments. 
Finally, we show the effect of introducing manual efforts to rectify incorrect labels (\cref{fig:ablation} With Manual). 
Our parsing method can also be deployed to annotate other existing 4D human datasets. We present examples of BUFF\cite{BUFF}, X-Humans \cite{X-Avatar}, and ActorsHQ\cite{HumanRF} and additional qualitative results in Supp. Mat.

\vspace{-.5em}
\section{Dataset Description}
\label{sec:dataset}
\vspace{-.5em}

{\datasetname} contains 520 motion sequences (150 frames at 30 fps) in 64 real-world human outfits with a total of 78k frames.
Each frame consists of multi-view images at 1k resolution, an 80k-face triangle 3D mesh with vertex annotations, and a 1k-resolution texture map.
We also provide each garment with its canonical template to benefit the clothing simulation study. 
Finally, each 3D scan is accurately registered by SMPL/SMPL-X body models.

To record {\datasetname} we recruited 32 participants (18 female), with an average age of 24. 
The dataset consists of 4 dresses, 30 upper, 28 lower, and 32 outer garments. 
Participants were instructed to perform different dynamic motions for each 5-second sequence. 
For each participant, we capture two types of outfits: {\bf Inner Outfit} comprising the inner layer dress/upper, and lower garments; 
and {\bf Outer Outfit} with an additional layer of garment, such as open jackets or coats. 
A unique feature of {\datasetname} is the challenging clothing deformations we captured. 
To quantify these deformations, we compute the mean distances from the garments to the registered SMPL body surfaces.
The inner and outer outfits exhibit distance ranges up to 7.12 cm and 14.76 cm over all frames.
This is twice as much as what we observed in the X-Humans dataset \cite{X-Avatar}, for example.
In the 10\% most challenging frames, this increases to 20.09 cm for outer outfits, highlighting the prevalence of challenging garments. 
Please refer to Supp. Mat. for dataset details.

\begin{figure*}[t]
\centering
\includegraphics[width=\linewidth]{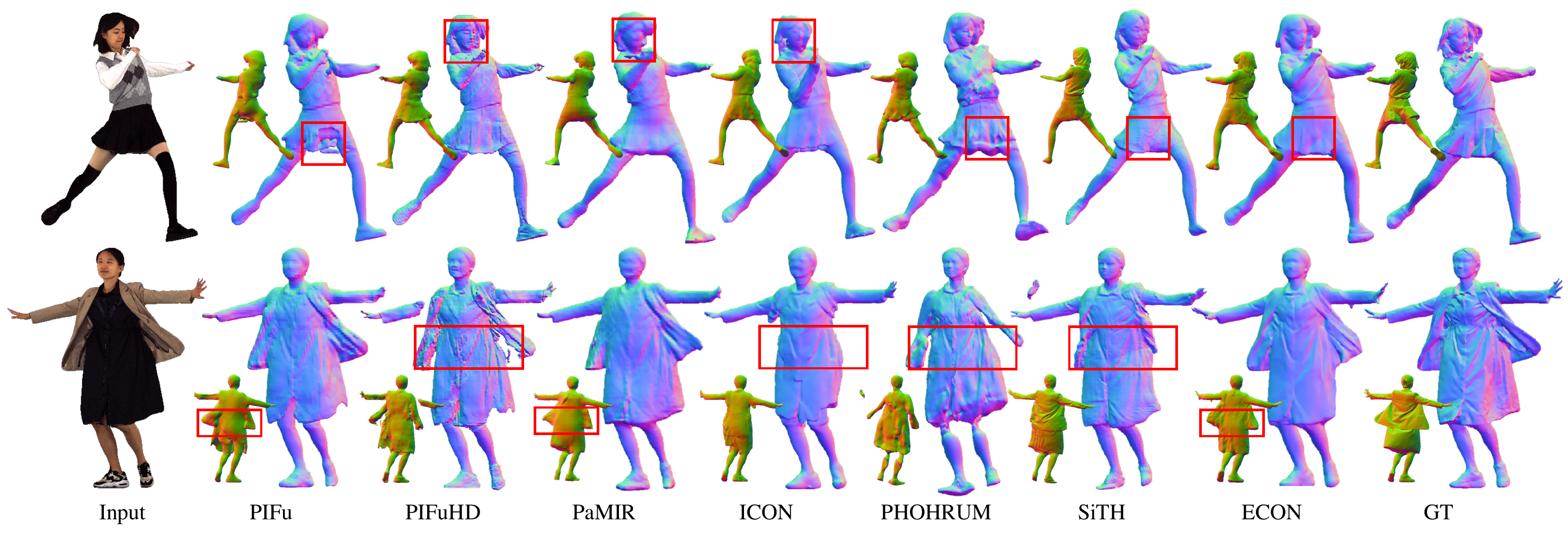}
\vspace{-1.2em}
\caption{\textbf{Examples of clothed human reconstruction on~\datasetname}. We evaluate state-of-the-art methods using both inner (\emph{Top}) and outer (\emph{Bottom}) outfits. We show that existing methods generally struggle with the challenging loose garments. Moreover, these approaches cannot faithfully recover realistic details such as clothing wrinkles.}
\label{fig:image_reco}
  \vspace{-1.5em}
\end{figure*}

\vspace{-0.5em}
\section{Benchmark Evaluation}
\label{Benchmark}
\vspace{-.5em}
With high-quality 4D scans and diverse garment meshes in dynamic motions, {\datasetname} serves as an ideal ground truth for a variety of computer vision and graphics benchmarks.
In our work, we outline several standard benchmarks conducted in these fields using our dataset.
Our primary focus is on tasks related to clothing simulation (\cref{sec:bench:garment_sim}) and clothed human reconstruction (\cref{sec:bench:cloth_recon}).
Additionally, benchmarks on human parsing and human representation learning are included in our Supp. Mat.

\newcommand{\myhs}{\hspace{3pt}}
\renewcommand{\arraystretch}{1.1}

\begin{table}[t]
    \centering
    \small
\resizebox{\columnwidth}{!}{%
\setlength\tabcolsep{3pt}
\begin{tabular}{l|c@{\myhs}c|c@{\myhs}c|c@{\myhs}c|c@{\myhs}c}
\toprule
    & \multicolumn{2}{c|}{Lower}  & \multicolumn{2}{c|}{Upper} & \multicolumn{2}{c|}{Dress}  & \multicolumn{2}{c}{Outer} \\
\midrule
    Method & CD $\downarrow$ & $E_{Str}$ $\downarrow$ & CD $\downarrow$ & $E_{Str}$ $\downarrow$  & CD $\downarrow$ & $E_{Str}$ $\downarrow$  & CD $\downarrow$ & $E_{Str}$ $\downarrow$ \\
\midrule
    LBS & 1.767 & 0.333 & 2.167 & 0.095 & 4.461 & 1.293 & 4.626 & 0.811\\
    PBNS~\cite{PBNS}  & 1.885 & 0.107 & 2.687 & 0.040 & 4.869 & 0.643 & 4.859 & 0.107 \\
    NCS~\cite{NCS} & 1.716 & 0.017 & 2.112 & 0.016 & 4.548 & 0.031 & 4.738 & 0.025 \\
    HOOD~\cite{HOOD} & 2.070 & 0.008 & 2.668 & 0.013 & 4.292 & 0.010 & 5.355 & 0.011 \\ \hline
    HOOD*  & 0.924 & 0.010 & 1.308 & 0.015 & 2.463 & 0.009 & 2.833 & 0.009\\

\bottomrule
\end{tabular}}
\vspace{-0.8em}
\caption{\textbf{Clothing simulation benchmark}. CD is Chamfer Distance between the simulation and ground truth. $E_{str}$ denotes stretching energy with respect to the template.}
  \vspace{-2em}
\label{tab:simulation}
\end{table}

\vspace{-0.5em}
\subsection{Clothing Simulation}
\label{sec:bench:garment_sim}
\vspace{-.5em}

\paragraph{Experimental setup.}
We introduce a new benchmark for clothing simulation, leveraging the garment meshes from {\datasetname}, which capture dynamical real-world clothing deformations. 
This benchmark evaluates three methods for modeling garment dynamics: \textbf{PBNS}~\cite{PBNS}, Neural Cloth Simulator (\textbf{NCS}~\cite{NCS}), and \textbf{HOOD}~\cite{HOOD}, as well as a baseline method that applies SMPL-based linear blend-skinning (\textbf{LBS}) to the template.
We ran the simulations using T-posed templates extracted from static scans and compared the results to the ground-truth garment meshes across various pose sequences.
Our evaluation metrics include the Chamfer Distance (\textbf{CD}), which compares the resulting mesh sequences with ground-truth point clouds, and the average stretching energy ($E_{str}$) calculated by measuring the difference in edge lengths between the simulated and template meshes.
The experiments were conducted across four categories of garments (Lower, Upper, Dress, and Outer), with four garment templates in each category. 
We simulated clothing deformation for each garment in six different pose sequences, providing a comprehensive comparison of their ability to generate realistic motions.

\vspace{-1.5em}
\paragraph{Fine-tuning material parameters.}
To demonstrate the advantages of real-world garment meshes in~\datasetname, we also introduce a simple optimization-based strategy for inverse simulation using HOOD. 
Specifically, we optimize the material parameters fed into the HOOD model to minimize the simulations' Chamfer Distance to the ground-truth sequences and their stretching energy. This optimized version is denoted as \textbf{HOOD*}.
For more details on the material optimization experiments, please refer to Supp. Mat. 

\vspace{-1.5em}
\paragraph{Evaluation results.}
The quantitative and qualitative comparisons of the clothing simulation methods are presented in \cref{tab:simulation} and \cref{fig:simulation} respectively.
The LBS baseline and LBS-based approaches (PBNS and NCS) perform better with upper and lower garments, which exhibit limited free-flowing motions compared with the dress and outer garments.
Conversely, HOOD excels with dresses, generating more natural, free-flowing motions and achieving lower stretching energy. 
However, if HOOD fails to generate realistic motions for a single frame, this error propagates to all subsequent frames.
This issue does not occur in the LBS-based methods, which generate geometries independently for each frame.
With finely-tuned material parameters, HOOD* produces garment sequences that more faithfully replicate real-world behavior. 
We anticipate that future research in learned garment simulation will increasingly focus on modeling real-world garments made from complex heterogeneous materials.
This will be a major step in creating realistically animated digital avatars, and we believe {\datasetname} will be highly instrumental in this task.

\begin{table}[t]
    \centering
    \small
\resizebox{\columnwidth}{!}{%
\begin{tabular}{l|ccc|ccc}
\toprule
                     & \multicolumn{3}{c|}{Inner}         & \multicolumn{3}{c}{Outer}          \\
\midrule
 Method               & CD$\downarrow$ & NC$\uparrow$ & IoU$\uparrow$ & CD$\downarrow$ & NC$\uparrow$ & IoU$\uparrow$      \\
\midrule
 PIFu~\cite{saito2019pifu}    & 2.696 & 0.792 & 0.690 &  2.783 & 0.759 & 0.697   \\
 PIFuHD~\cite{saito2020pifuhd}  & \underline{2.426} & 0.793 & 0.739 & \underline{2.393} & 0.763 & 0.743  \\
 PaMIR~\cite{zheng2021pamir}   & 2.520 & \underline{0.805} & 0.706 &  2.608 & \underline{0.777} & 0.715   \\
 ICON~\cite{xiu2022icon}    & 2.473 & 0.798 & \underline{0.752} &  2.832 & 0.762 & \textbf{0.756}  \\
 PHORHUM~\cite{alldieck2022phorhum} & 3.944 & 0.725 & 0.580 &  3.762 & 0.705 & 0.603  \\
 ECON~\cite{xiu2023econ}    & 2.543 & 0.796 & 0.736 &  2.852 & 0.760 & 0.728 \\
 SiTH~\cite{ho2024SiTH}     & \textbf{2.110} & \textbf{0.824} & \textbf{0.755} &  \textbf{2.322} & \textbf{0.794} & \underline{0.749} \\
\bottomrule
\end{tabular}
}
\vspace{-1em}
\caption{\textbf{Clothed human reconstruction benchmark}. We computed Chamfer distance (CD), normal consistency (NC), and Intersection over Union (IoU) between ground truth and reconstructed meshes obtained from different baselines.}
  \vspace{-2em}
\label{tab:image_reco}
\end{table}

\vspace{-.5em}
\subsection{Clothed Human Reconstruction}
\label{sec:bench:cloth_recon}
\vspace{-.5em}
\paragraph{Experimental setup.}
We create a new benchmark for evaluating state-of-the-art clothed human reconstruction methods on the ~\datasetname~dataset. This benchmark is divided into three subtasks. First, we evaluate \textbf{single-view human reconstruction} utilizing images and high-quality 3D scans from our dataset. In addition, benefiting from the garment meshes in our dataset, we establish the first real-world benchmark for evaluating \textbf{single-view clothing reconstruction}. Finally,  we assess \textbf{video-based human reconstruction} approaches leveraging the sequences in \datasetname~ that capture rich motion dynamics of both human bodies and garments.
In all the experiments, we report 3D metrics including Chamfer Distance (\textbf{CD}), Normal Consistency (\textbf{NC}), and Intersection over Union (\textbf{IoU}) to compare the predictions with ground-truth meshes.
\vspace{-1.5em}
\paragraph{Single-view human reconstruction.}
We use the two test sets defined in~\cref{sec:dataset} (denote as \textbf{Outer} and \textbf{Inner}) to evaluate the following single-view reconstruction methods: \textbf{PIFu}~\cite{saito2019pifu}, \textbf{PIFuHD}~\cite{saito2020pifuhd}, \textbf{PaMIR}~\cite{zheng2021pamir}, \textbf{ICON}~\cite{xiu2022icon}, \textbf{PHORHUM}~\cite{alldieck2022phorhum}, \textbf{ECON}~\cite{xiu2023econ}, and \textbf{SiTH}~\cite{ho2024SiTH}. The evaluation results are summarized in~\cref{fig:image_reco} and~\cref{tab:image_reco}. We observed that methods leveraging SMPL body models as guidance (i.e., ICON, ECON, SiTH) performed better in reconstructing inner clothing. However, their performance significantly declined when dealing with outer garments. On the other hand, end-to-end models like PIFu and PIFuHD demonstrated more stability with both clothing types. This leads to an intriguing research question: whether the human body prior is necessary for reconstructing clothing. Qualitatively, we see that even the best-performing methods cannot perfectly reconstruct realistic free-flowing jackets as shown in~\cref{tab:image_reco}. We believe~\datasetname~will offer more valuable insights for research in clothed human reconstruction. 
\begin{figure}[t]
\centering
\includegraphics[width=\linewidth]{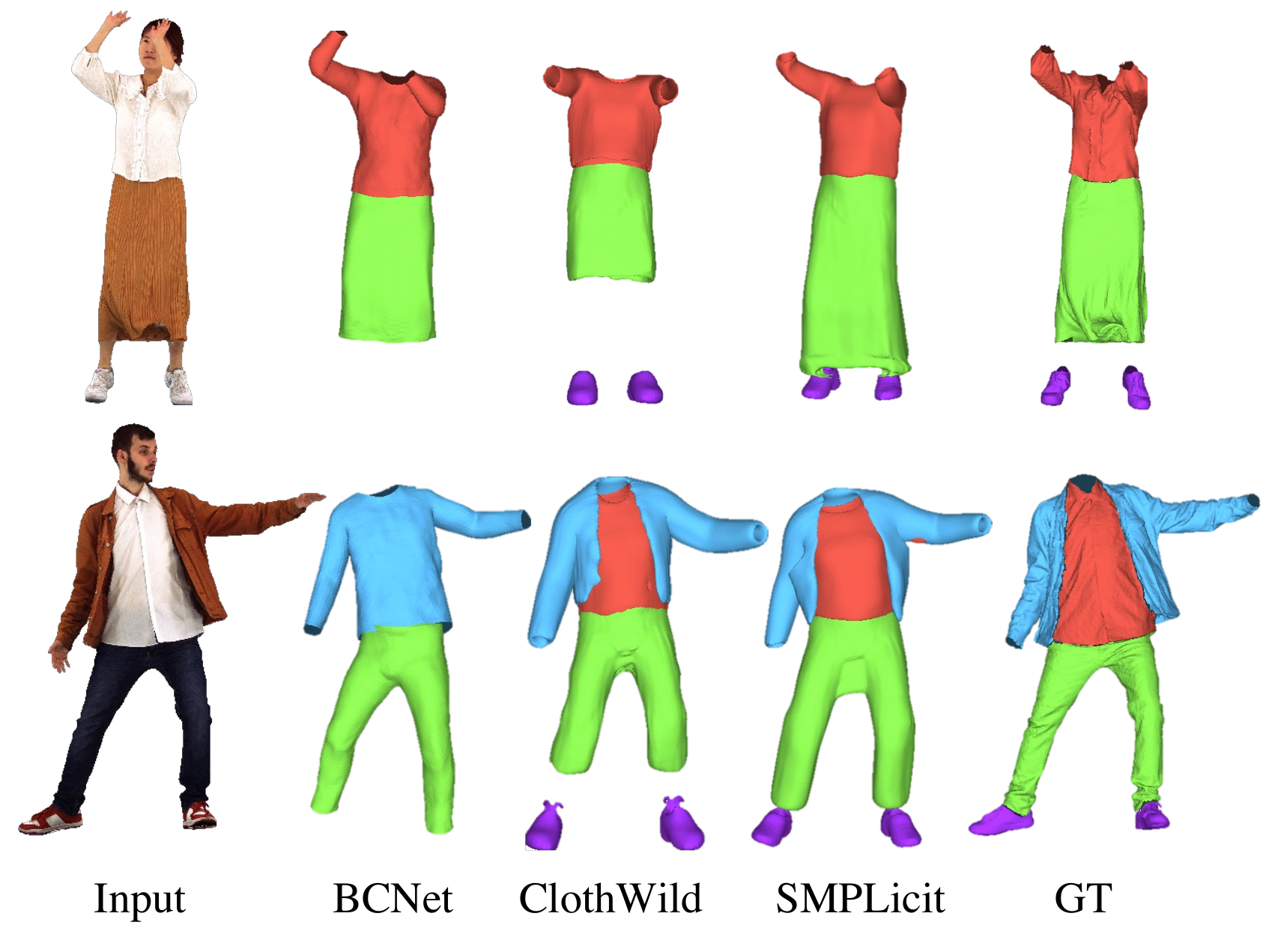}
  \vspace{-1.2em}
\caption{\textbf{Examples of clothing reconstruction on~\datasetname}. We visualize the reconstructed garment meshes from different approaches. These methods trained on synthetic datasets failed to predict accurate clothing sizes and detailed wrinkles.}
\label{fig:cloth_reco}
  \vspace{-1.em}
\end{figure}
\renewcommand{\arraystretch}{1.1}

\begin{table}[t]
    \centering
    \small
\resizebox{\columnwidth}{!}{%
\setlength\tabcolsep{3pt}
\begin{tabular}{l|c@{\myhs}c|c@{\myhs}c|c@{\myhs}c|c@{\myhs}c}
\toprule
    & \multicolumn{2}{c|}{Shoes}  & \multicolumn{2}{c|}{Lower} & \multicolumn{2}{c|}{Upper}  & \multicolumn{2}{c}{Outer} \\
\midrule
    Method & CD $\downarrow$ & IoU$\uparrow$ & CD $\downarrow$ & IoU$\uparrow$  & CD $\downarrow$ & IoU$\uparrow$  & CD $\downarrow$ & IoU$\uparrow$ \\
\midrule
    BCNet~\cite{jiang2020bcnet}         & - & - & 2.533 & 0.675 & 2.079 & 0.700 & 3.600 & 0.639 \\
    SMPLicit~\cite{corona2021smplicit}  & 2.619 & 0.621 & 2.101 & 0.698 & 2.452 & 0.617 & 3.359 & 0.618 \\
    ClothWild~\cite{Moon_2022_ECCV_ClothWild} & 3.657 & 0.548 & 2.690 & 0.582 & 3.279 & 0.533 & 4.163 & 0.588  \\
\bottomrule
\end{tabular}}
\vspace{-1.em}
\caption{\textbf{Clothing reconstruction benchmark}. We report Chamfer Distance (CD), and Intersection over Union (IoU) between the ground-truth garment meshes and the reconstructed clothing.}
\label{tab:cloth_recon}
\vspace{-2.em}
\end{table}

\vspace{-1.6em}
\paragraph{Single-view clothes reconstruction.} Clothes reconstruction has received relatively little attention compared to full-body human reconstruction.
Leveraging the garment meshes in~\datasetname, we introduce the first real-world benchmark to assess prior art, including \textbf{BCNet}~\cite{jiang2020bcnet}, \textbf{SMPLicit}~\cite{corona2021smplicit}, and \textbf{ClothWild}~\cite{Moon_2022_ECCV_ClothWild}. 
The results of different clothing types, as shown in~\cref{fig:cloth_reco}, indicate a significant gap between the reconstructed and real clothing.
Firstly, the clothing sizes produced by these methods are often inaccurate, suggesting a lack of effective use of image information for guidance.
Moreover, the results typically lack geometric details like clothing wrinkles compared to full-body reconstruction.
We report quantitative results in~\cref{tab:cloth_recon}.
We observed that the data-driven method (BCNet) performs better with inner clothing, while the generative fitting method (SMPLicit) shows more robustness to outer clothing, such as coats. 
However, none of these methods is designed for or trained on real-world data. The domain gap between synthetic and real data still limits their capability to produce accurate shapes and fine-grained details.
We expect our benchmark and dataset will draw more research attention to the topic of real-world clothing reconstruction.

\vspace{-1.em}

\begin{figure}[t]
\centering
\vspace{-1em}
\includegraphics[width=\linewidth]{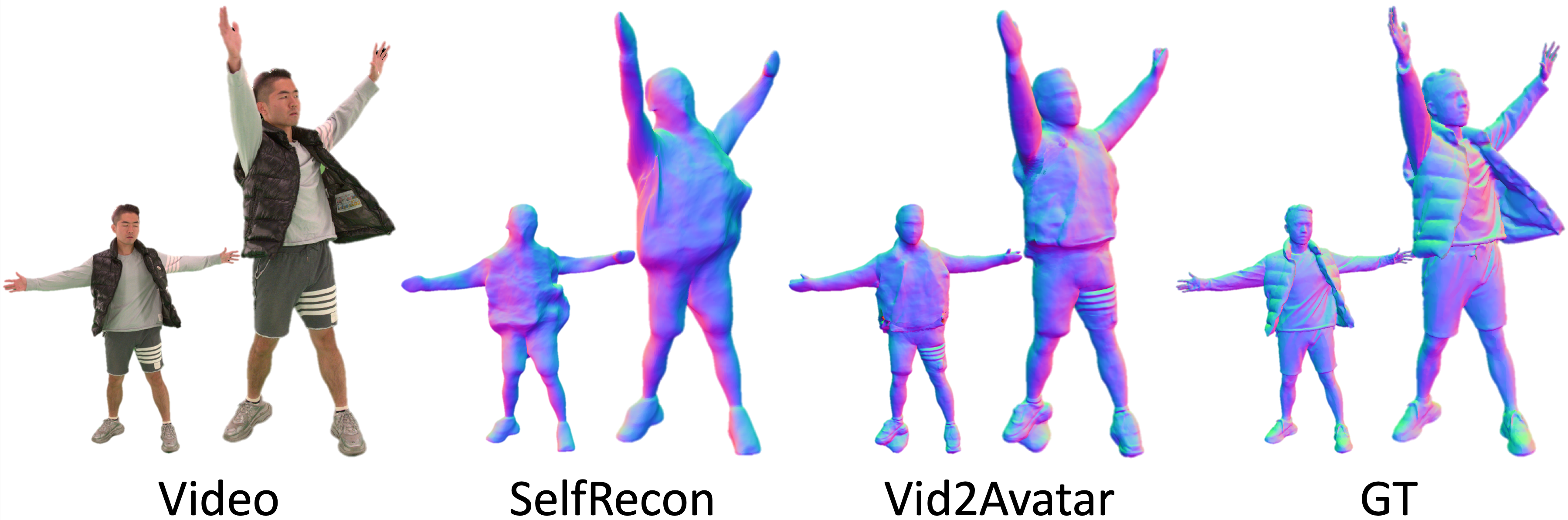}
\vspace{-.5em}
\caption{\textbf{Video-based human reconstruction.} Qualitative results of video-based human reconstruction methods on~\datasetname. Prior works struggle to reconstruct 3D human with challenging outfits and cannot recover the fine-grained surface details.}
\label{fig:videorecon}
   \vspace{-1.em}
\end{figure}

\paragraph{Video-based human reconstruction} 
Leveraging the sequential 4D data in our dataset, we create a new benchmark for evaluating video-based human reconstruction methods. 
We applied \textbf{Vid2Avatar}~\cite{Vid2Avatar} and \textbf{SelfRecon}~\cite{SelfRecon} to obtain 4D reconstructions and compared them with the provided ground-truth 4D scans.
As observed in~\cref{fig:videorecon}, both methods struggle with diverse clothing styles and face challenges in reconstructing surface parts that greatly differ in topology from the human body, such as the open jacket.
Moreover, there remains a noticeable discrepancy between the real geometry and the recovered surface details. 
Quantitatively, the existing methods cannot achieve satisfactory reconstruction results with outer garments, as demonstrated by a large performance degradation in~\cref{tab:videorecon}.
We believe~\datasetname~provides essential data for advancing video-based human reconstruction methods, particularly in achieving detailed geometry recovery for challenging clothing.

\begin{table}[t]
    \centering
    \small
\resizebox{\columnwidth}{!}{%
\begin{tabular}{l|ccc|ccc}
\toprule
                  & \multicolumn{3}{c|}{Inner} &  
                    \multicolumn{3}{c}{Outer} \\
\midrule
Method & CD$\downarrow$ & NC$\uparrow$ & IoU$\uparrow$ & CD$\downarrow$ & NC$\uparrow$ & IoU$\uparrow$   \\
 \midrule
 SelfRecon~\cite{SelfRecon}      & 3.180 & 0.729 & 0.754 & 4.027  & 0.683  & 0.745 \\
 Vid2Avatar~\cite{Vid2Avatar}    & 2.870 & 0.750 & 0.772 & 3.014  & 0.725  & 0.787 \\
\bottomrule
\end{tabular}}
\caption{\textbf{Video-based human reconstruction}. Results of video-based human reconstruction methods on~\datasetname.}
\vspace{-1em}
\label{tab:videorecon}
\end{table}
\begin{table}[t]
    \centering
    \small
\begin{tabular}{l|cc|cc}
\toprule
& \multicolumn{2}{c|}{Inner} & \multicolumn{2}{c}{Outer} \\
\midrule
Method & mAcc.$\uparrow$ & mIoU$\uparrow$  & mAcc.$\uparrow$ & mIoU$\uparrow$  \\
\midrule
SCHP~\cite{SCHP}    & 0.908 & 0.832 & 0.863 & 0.768 \\
CDGNet~\cite{CDGNet}  & 0.922 & 0.853 & 0.887 & 0.790 \\
Graphonomy~\cite{Graphonomy} & 0.968 & 0.859 & 0.915 & 0.810  \\
\bottomrule
\end{tabular}
\caption{\textbf{Image-based human parsing}. Results of image-based human parsers on \datasetname.}
\vspace{-1em}
\label{tab:parsing}
\end{table}

\begin{table}[t]
    \centering
    \small
\resizebox{\columnwidth}{!}{%
\begin{tabular}{l|ccc|ccc}
\toprule
& \multicolumn{3}{c|}{Inner} &   \multicolumn{3}{c}{Outer} \\
\midrule
Method & CD$\downarrow$ & NC$\uparrow$ & IoU$\uparrow$ & CD$\downarrow$ & NC$\uparrow$ & IoU$\uparrow$   \\
 \midrule
 SCANimate~\cite{SCANimate}           & 0.965 & 0.854 & 0.918 & 1.237  & 0.828  & 0.912 \\
 SNARF~\cite{SNARF}          & 1.158 & 0.843 & 0.907 & 1.248 & 0.827 & 0.930  \\
 X-Avatar~\cite{X-Avatar}    & 1.008 & 0.861 & 0.954 & 1.177 & 0.841 & 0.946  \\
\bottomrule
\end{tabular}}
\caption{\textbf{Human representation learning}. Results of human representation learning approaches on~\datasetname.}
\vspace{-2em}
\label{tab:avatar}
\end{table}

\subsection{Clothed Human Parsing}
\label{sec:bench:image_parsing}
\begin{figure*}[t]
\centering
\includegraphics[width=0.9\linewidth]{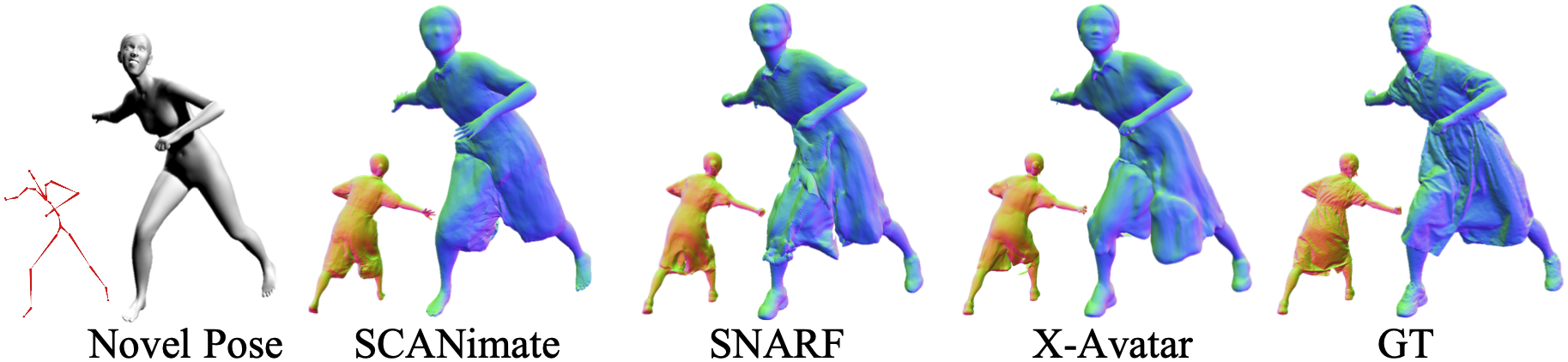}
\caption{\textbf{Human representation learning.} Qualitative results of the novel pose synthesis of state-of-the-art human representation learning approaches together with the GT of~\datasetname. All Baseline methods fail to learn the large non-rigid surface deformations and are bounded by the skeletal deformations.}

\label{fig:avatar}
\end{figure*}

\begin{figure}[t]
\centering
\includegraphics[width=1\linewidth]{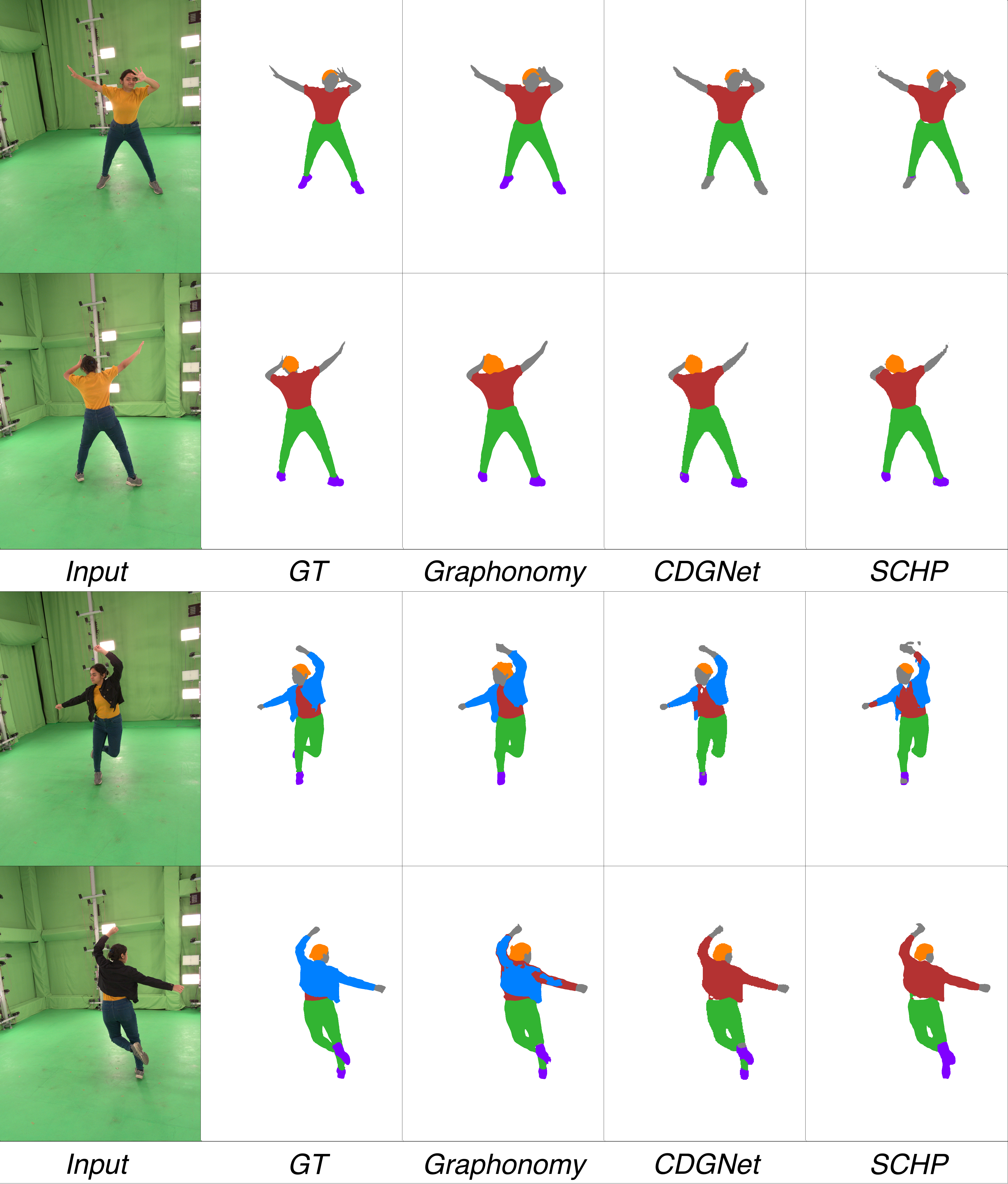}
\vspace{-0.6em}
\caption{\textbf{Human parsing comparison.} We use the ground-truth semantic labels to evaluate state-of-the-art human parsing models. These methods generally failed to predict correct clothing labels from different view angles.}
\vspace{-2em}
\label{fig:supp_parser}
\end{figure}

We design a benchmark for the image-based human parser. Concretely, we project each scan frame's vertex labels to the multi-view captured images using corresponding camera parameters and rasterizer, which provide the ground-truth pixel labels for evaluating the image-based human parsing methods: \textbf{SCHP} \cite{SCHP}, \textbf{CDGNet} \cite{CDGNet}, and \textbf{Graphonomy} \cite{Graphonomy}. In~\cref{tab:parsing}, we report the mean Pixel Accuracy ($\bf mAcc. $) and mean Intersection over Union ($ \bf mIOU $) between the prediction and the ground-truth labels. We conducted our human image parsing experiments on one subset of our 4D-DRESS dataset, which contains 128 sequences of 64 outfits (2 sequences for each of the inner and outer outfits). The qualitative parsing results are shown in \cref{fig:supp_parser}.

\subsection{Human Representation Learning}
\label{sec:bench:human_learning}

We design a new benchmark for evaluating the human representation learning task. Unlike physics-based methods, this line of work directly takes 3D human scans as training input and obtains an animation-ready human avatar. We follow the split strategy mentioned before and evaluate prior works, \textbf{SCANimate} \cite{SCANimate}, \textbf{SNARF} \cite{SNARF}, \textbf{X-Avatar} \cite{X-Avatar} on the novel-pose synthesis. \cref{fig:avatar} shows that state-of-the-art human representation learning approaches cannot correctly learn the large non-rigid surface deformations (e.g., folded skirt) due to the strong skeletal dependency and the lack of modeling for temporal dynamics. This effect can also be reflected in \cref{tab:avatar} quantitatively where all baseline methods produce higher errors on the split of more challenging garments (outer outfits).

\section{Discussion}
\textbf{Limitations.} Our current pipeline requires substantial computational time. The offline manual rectification process and garment mesh extraction also demand expertise in 3D editing and additional human efforts. These factors constrain the scalability of our dataset. With a goal of expanding more diverse subjects and clothing, real-time 4D annotation and rectification/editing will be exciting future work.\\
\textbf{Conclusion.}~\datasetname~is the first real-world 4D clothed human dataset with semantic annotations, aiming to bridge the gap between existing clothing algorithms and real-world human clothing. 
We demonstrate that \datasetname~is not only a novel data source but also a challenging benchmark for clothing simulation, reconstruction, and other related tasks. We believe that \datasetname~can support a wide range of endeavors and foster research progress by providing high-quality 4D data in life like human clothing. \\
\textbf{Acknowledgements.} This work was partially supported by the Swiss SERI Consolidation Grant "AI-PERCEIVE". AG was supported in part by the Max Planck ETH CLS.

{
    \small
    \bibliographystyle{ieeenat_fullname}
    \bibliography{main}
}

\clearpage

\clearpage
\maketitlesupplementary

\section{Implementation Details}

\begin{figure}[t]
\centering
\includegraphics[width=1.\linewidth]{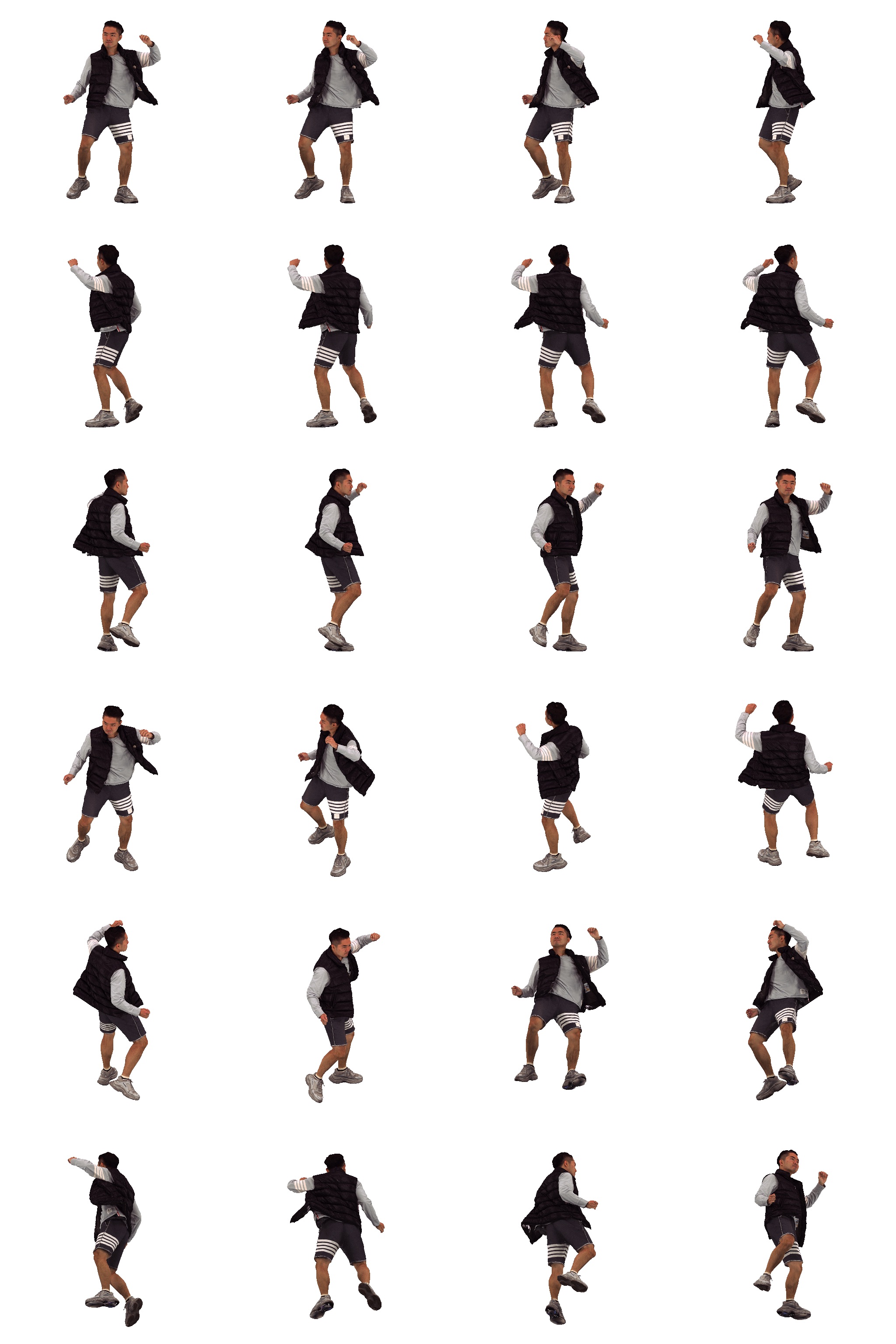}
\vspace{-0.6em}
\caption{\textbf{Example of 24 rendered views.} We render 24 views to ensure the visibility of each scan vertex and consider the computational cost of human parsing.}
\label{fig:supp_render}
  \vspace{-1.em}
\end{figure}

\begin{figure*}[t]
\centering
\includegraphics[width=1\linewidth]{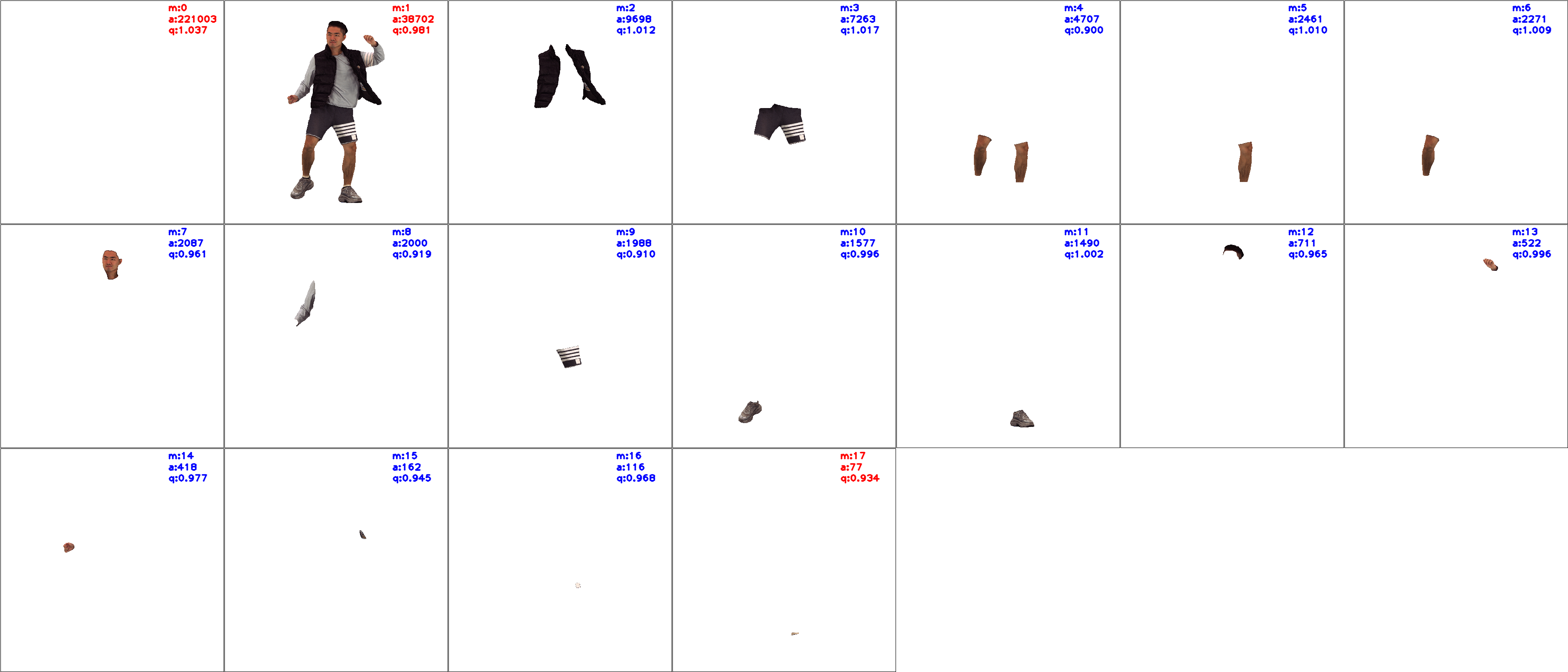}
\vspace{-0.6em}
\caption{\textbf{Example of SAM predictions.} The input image is the first view (upper-left) of \cref{fig:supp_render}. We filter out the segmentation masks that contain background, full body, and only small regions (marked as red).}
\label{fig:supp_mask}
\end{figure*}

\subsection{Multi-view Parsing}

\paragraph{Multi-view rendering.}
For each frame $ k \in \left\{ 1,...,{N}_{frame}\right\} $, we render twelve horizontal, six upper, and six lower images $ I_{img, n, k} $ that are uniformly distributed on a sphere by rasterizing the textured scan with Pytorch3D~\cite{Pytorch3D}, where $ n \in \left\{ 1,...,{N}_{view}=24\right\} $. Each scan is centralized according to its bounding box center and then placed at the camera sphere center. The rendered images have a resolution of $ 512 \times 512$. Examples of 24-view rendered images are shown in \cref{fig:supp_render}.

\renewcommand{\arraystretch}{1.2}
\begin{table}[t]
    \centering
\begin{tabular}{l|c}
\toprule
{\bf 4D-DRESS} & {\bf Graphonomy (LIP)} \\ \hline
(-1) other & background \\ \hline
\multirow{2}{*}{(0) skin} &  torso-skin, face, glove \\ & left-arm, right-arm, left-leg, right-leg \\ \hline
(1) hair & hat, hair, sunglasses \\ \hline
(2) shoe & socks, left-shoe, right-shoe \\ \hline
(3) upper & upper-clothing, dress, scarf \\ \hline
(4) lower & pant, skirt\\ \hline
(5) outer & coat\\
\bottomrule
\end{tabular}
\caption{\textbf{Label mapping between 4D-DRESS and LIP dataset}. We define 6 label categories based on LIP dataset.}
\label{tab:labeling}
\end{table}

\paragraph{Human image parser (\emph{PAR}).}
We apply the pre-trained Graphonomy~\cite{Graphonomy} to each rendered image $ I_{img, n, k} $ and save the label results as a new image $ I_{par, n, k} $. Concretely, we manually classify the 20 classes of Graphonomy labels into 6 classes that are used in our dataset: skin (0), hair(1), shoes(2), upper(3), lower(4), and outer(5) clothes. The corresponding labels between Graphonomy (LIP) and ours are shown in \cref{tab:labeling}. Specifically, we map the background label from Graphonomy to our setting with a label value -1, and the color code of white. These background labels will return 0 in the vote function $ f_{par, n}(p, l) $.

\paragraph{Optical flow transfer (\emph{OPT}).}
To establish connections with previous frames, we project previous frame vertex labels to multi-view labels $ I_{lab, n, k-1} $ using the same rendering cameras and rasterizer from Pytorch3D. Then, we warp these previous multi-view labels to the current frame $ I_{opt, n, k} $ using the optical flow vectors predicted by the RAFT \cite{RAFT} model. The vertex labels at the first frame do not involve this process thanks to our first-frame initialization (see ~\cref{supsec:manual}). Concretely, each pixel label with location $ p $ within $ I_{lab, n, k-1} $ will be warped to a new pixel location $ p + v $ at the current frame, through the optical flow vector $ v = RAFT(I_{img, n, k-1}, I_{img, n, k}, p) $. The new labels at the current frame are determined by voting. If there is no corresponding label found in the previous frame, the new label will be set to -1. 

\paragraph{Segmentation masks and scores (\emph{SAM}).}
We use Segment Anything Model \cite{SAM} to segment each rendered image $ I_{img, n} $ into a group of masks $ M_{m, n} $ without any extra prompts, where $ m \in \left\{ 1,...,{M}_{mask, n}\right\}$. Then we compute the score function $ S(l, M_{m, n}) $ within each mask for each label by fusing the votes from the image parser and optical flow, normalized by the area of the mask. \cref{fig:supp_mask} depicts the predicted segmentation masks from a rendered image. A pixel $ p $ within the rendered image $ I_{img, n} $ may belong to multiple segmentation masks. In this case, the SAM vote function $ f_{sam, n}(p, l) $ is calculated by summing all the scores of masks that contain this pixel.

\subsection{Graph Cut Optimization}

The energy Eq. (5) in the main paper is optimized through the graph cut algorithm (alpha-expansion). The vertex-wise unary energy is normalized among all labels and then added to the edge-wise binary energy. The weights are empirically set as $ \lambda_{p}=0.5 $, $ \lambda_{o}=0.5 $, $ \lambda_{po}=1.5 $, $ \lambda_{s}=1 $, and $ \lambda_{b}=1 $.

\begin{figure}[t]
\centering
\includegraphics[width=1.\linewidth]{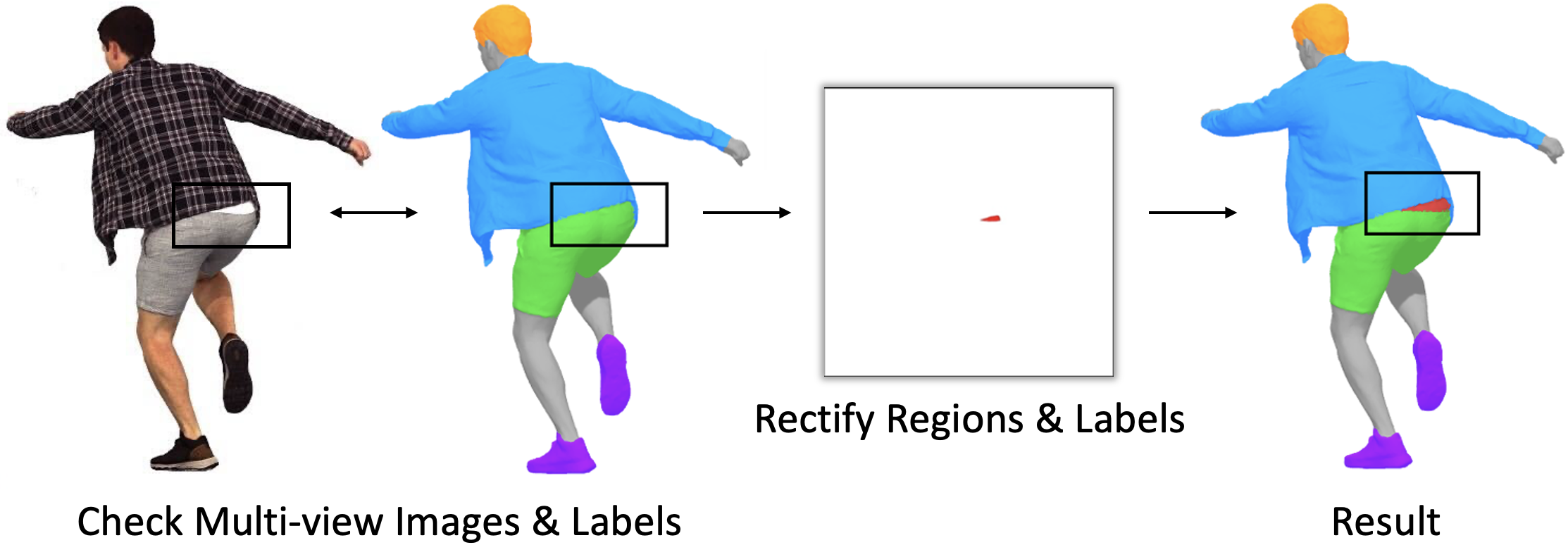}
\vspace{-0.6em}
\caption{\textbf{Example of manual rectification.} An annotator selects a region in the rendered images and gives a correct label. The label is projected to 3D and used for correcting the 3D vertices through a second round of graph cut optimization. }
\vspace{-0.6em}
\label{fig:supp_manual}
\end{figure}

\begin{figure*}[t]
\centering
\includegraphics[width=1\linewidth]{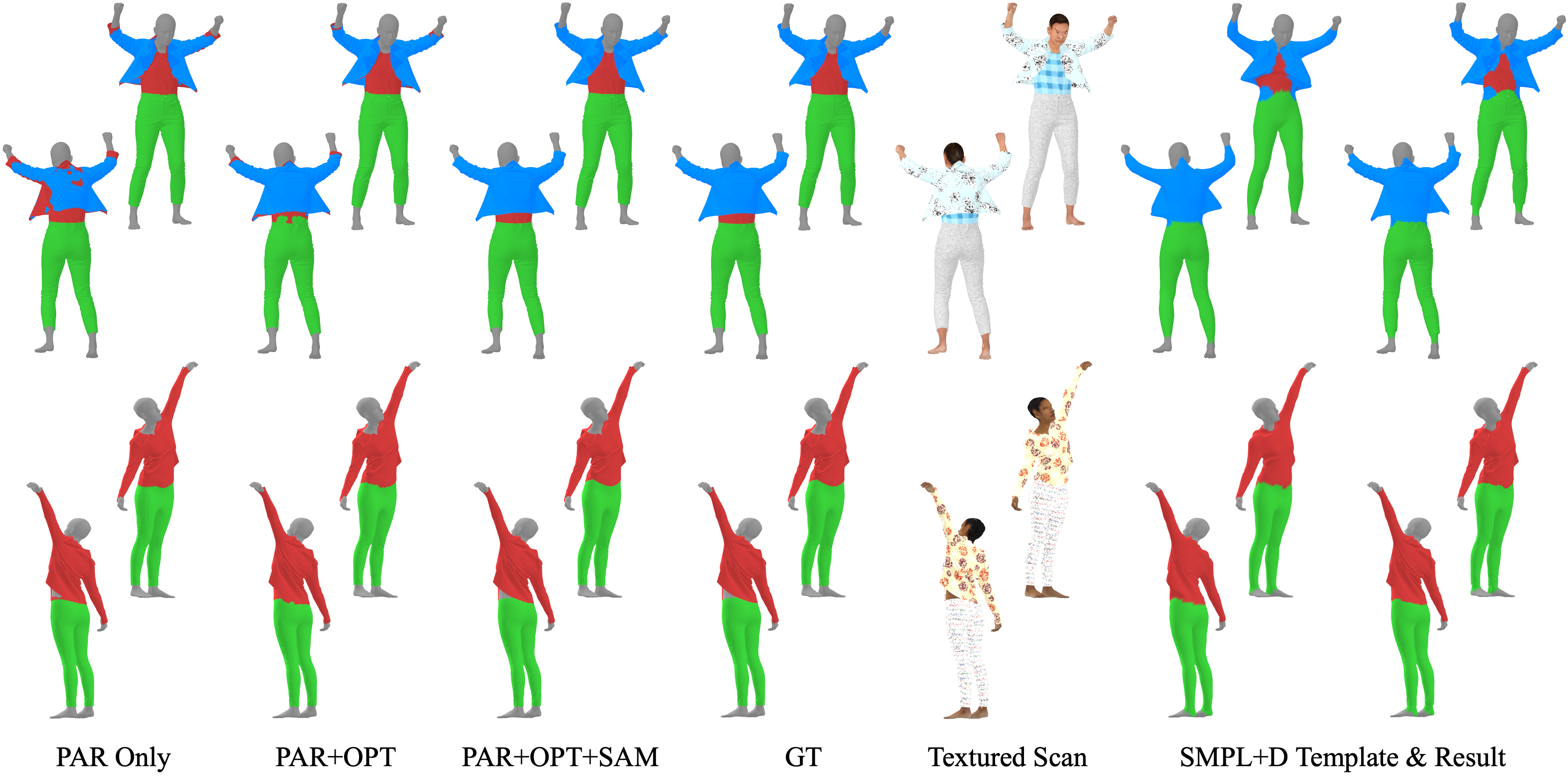}
\vspace{-0.6em}
\caption{\textbf{Ablation study and baseline comparison on the BEDLAM dataset.} We conducted ablative experiments on the synthetic BEDLAM dataset where ground-truth semantic labels are available.}
\label{fig:supp_experiment}
  \vspace{-1.em}
\end{figure*}

\subsection{Manual Rectification Process}
\label{supsec:manual}
\paragraph{Manual rectification on segmentation masks.}
In our dataset, each scan mesh has around 80k vertices.
Manually annotating their vertex labels on the 3D scans is very expensive and time-consuming.
Thus, we introduce a manual rectification process within the 2D image space.
After the first graph cut optimization, we render vertex labels to multi-view images, from which we let an annotator correct labels with the segmentation masks and a painting tool.
More specifically, the annotator is asked to identify an incorrectly labeled region by checking the multi-view images and labels.
Once an incorrect labeling is found, the annotator will look for its corresponding segmentation masks for label correction.
If such a mask does not exist, the annotator will manually paint the region using a painting tool. 
Finally, the images with rectified labels are projected to 3D vertices and are formulated as the manual vote function $ f_{man, n}(p, l) $.
The energy $ E_{man, n} $ term will be added to the second round of graph cut optimization, with a large weight $ w_{man} = 10 $. 
We note that for each 150-frame 4D sequence, the rectification process takes about 30 minutes on a desktop with an RTX 2080Ti GPU whereas the human parsing and the graph cut optimization take two and one hour, respectively.
An example of our rectification process is shown in \cref{fig:supp_manual}.

\paragraph{First-frame initialization of vertex labels.}
To ensure a good label initialization, the motion sequences always start from the A pose, which is easier for human parsing and pose registration. We obtain the first-frame vertex labels using the edge-wise binary energy and the multi-view unary energy calculated only from the image parser ($ E_{par} $) and manual rectifications ($ E_{man} $).

\section{Additional Parsing Experiments}
\begin{figure*}[t]
\centering
\includegraphics[width=1\linewidth]{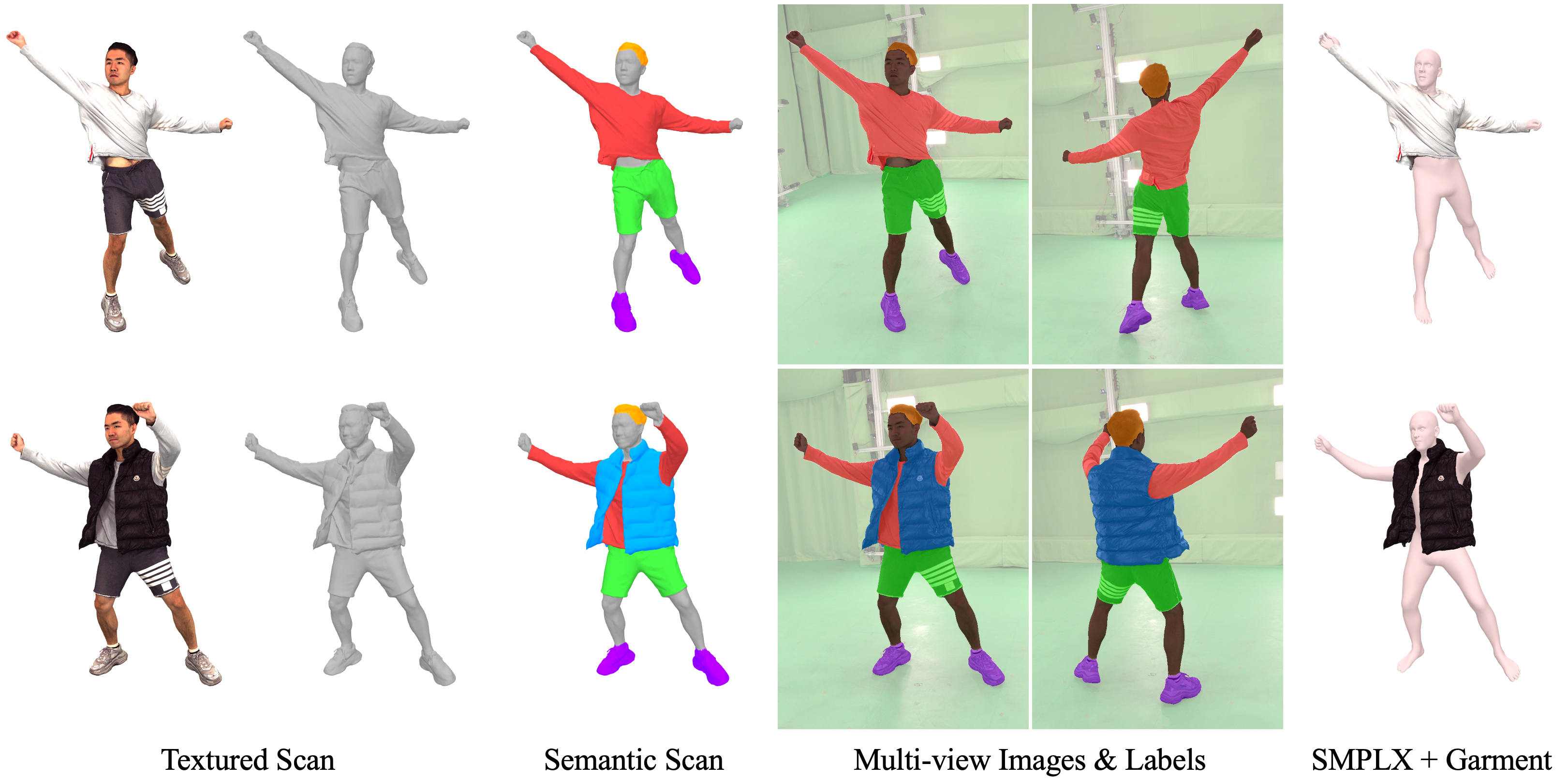}
\vspace{-0.6em}
\caption{\textbf{Data provided in the 4D-DRESS dataset.} We provide high-quality 4D textured scans. For each scan, we annotate vertex-level semantic
labels, thereby obtaining the corresponding garment meshes and fitted SMPL(-X) body meshes.}
\label{fig:supp_dataset}
  \vspace{-1.em}
\end{figure*}

\subsection{4D Parsing on Synthetic Datasets}

We conducted controlled 4D parsing experiments on two synthetic datasets, CLOTH4D \cite{CLOTH4D} and BEDLAM \cite{BEDLAM}, where the cloth meshes are simulated from cloth templates on top of the parameterized body models. Since within these synthetic datasets, some inner body and cloth vertices are always invisible from the outside, we report our labeling accuracy only on the vertices that are visible from our 24 views of rendered images.

\paragraph{Baseline comparison.} 
We first compare our 4D human parsing method with a template-based baseline method \cite{ClothCap} that utilizes a semantic SMPL+D template to first track the clothed human shape, and then project the template labels to neighboring scan vertices. Since ClothCap \cite{ClothCap} didn't release their 4D parsing code, we implemented their parsing method following their descriptions. We first register the SMPL+D model to all frames. Then we initialize the first frame template label using the nearby scan vertex labels obtained through our first-frame initialization process. At each frame, we update the template labels using the body prior, previous frame prior, and the Gaussian Mixture Model trained from the vertex colors of each labeled category. Finally, the scan vertex labels are assigned from the nearest template label. The quantitative parsing results from this baseline method are shown in the main paper. Here, we show more qualitative results in \cref{fig:supp_experiment}.

The main issue of this template-based baseline method is fitting the SMPL+D template to loose human outfits. The spatial mismatch between template and loose garments generates incorrect labels, especially in the open area of the jackets. Besides this, precisely updating the template labels using the Gaussian mixture model of labeled vertex colors is also difficult, especially in front of garments that have similar colors. The limited template resolution also results in noisy boundary labels at the higher-resolution clothed human meshes. The parsing accuracy from this baseline method is below 90\% for all synthetic outfits.

\paragraph{Ablation studies.} 
We then compare our 4D human parsing method (without manual rectifications) with several ablations of the multi-view parsing inputs (PAR Only, PAR+OPT, PAR+OPT+SAM), as shown in \cref{fig:supp_experiment}. Similar to Fig. 3 in the main paper, we observed similar qualitative results on the synthetic datasets.

\begin{figure}[t]
\centering
\includegraphics[width=0.85\linewidth]{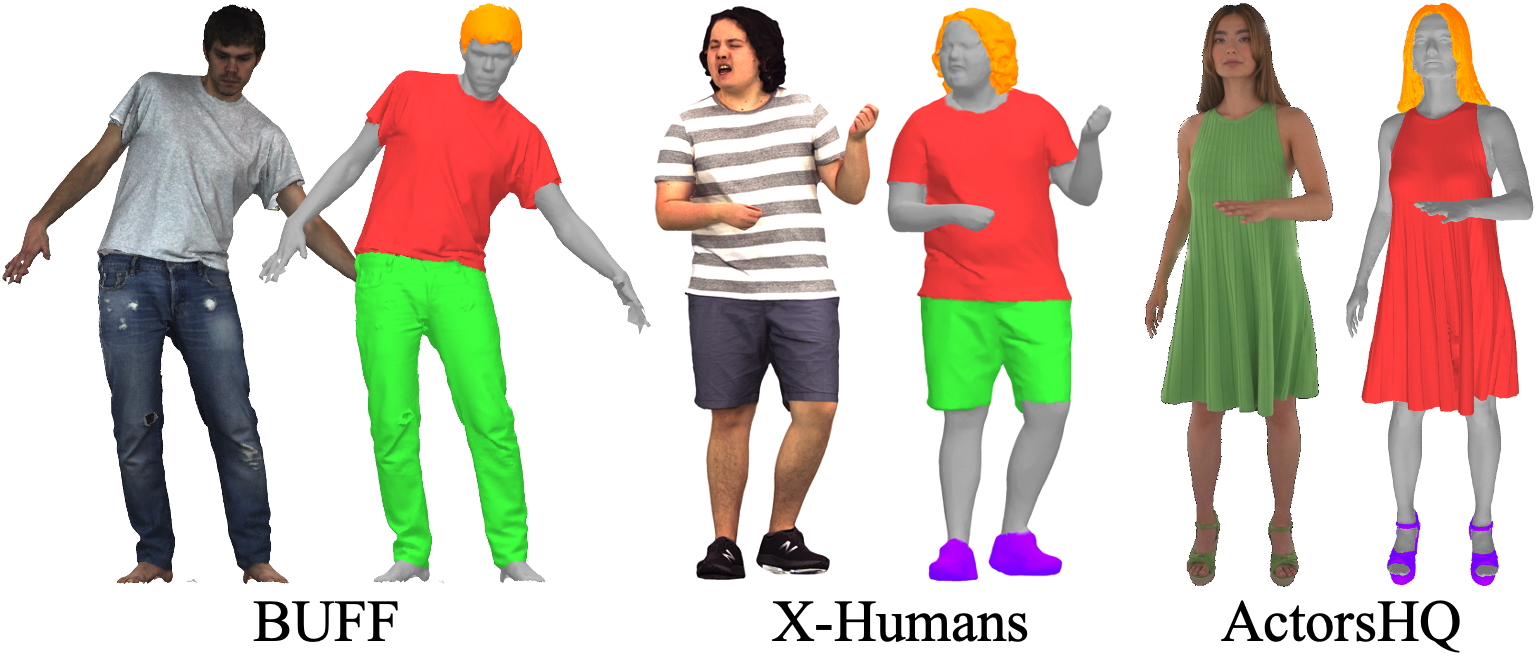}
\vspace{-1em}
\caption{\textbf{4D human parsing on other real-world datasets.}}
\label{fig:rebuttal_fig1}
  \vspace{-1.em}
\end{figure}

\subsection{4D Parsing on Other Datasets}
Our 4D human parsing method takes the input as scan mesh sequences and multi-view videos and thus can be applied to the existing real-world 4D human datasets, such as BUFF, X-Humans, and ActorsHQ, as shown in \cref{fig:rebuttal_fig1}.

\begin{figure}[t]
\centering
\includegraphics[width=0.8\linewidth]{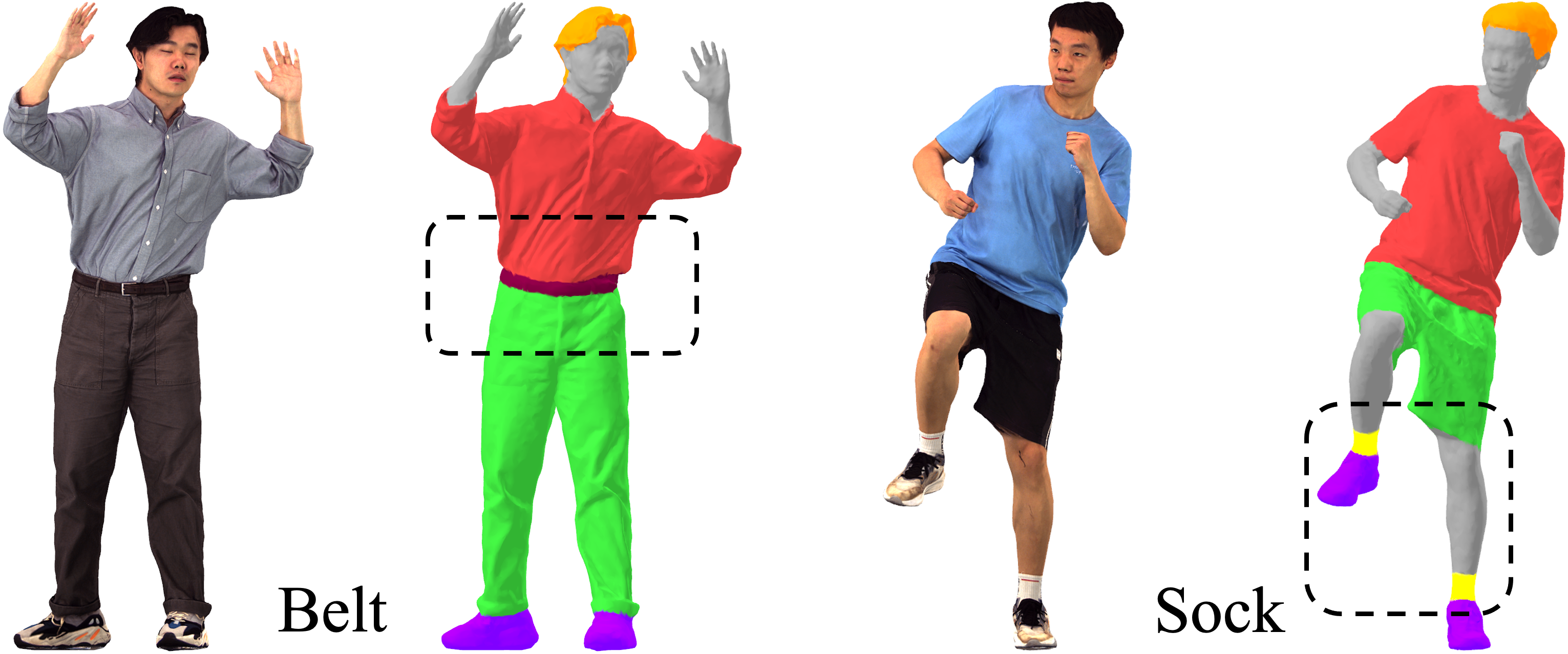}
\vspace{-1em}
\caption{\textbf{4D human parsing with new labels.}}
\label{fig:rebuttal_fig2}
  \vspace{-1.em}
\end{figure}

\subsection{4D Parsing with New Labels}
The six classes in our 4D-DRESS are strategically defined to ensure a consistent benchmark evaluation for clothing simulation and reconstruction. We showcase the generalization ability of our parsing method with new labels in \cref{fig:rebuttal_fig2}, by effectively distinguishing a belt from pants and socks from shoes. Initiated during the first-frame initialization, these new labels can integrate into the 4D parsing pipeline. However, refining labels for these smaller clothes and objects may entail additional manual efforts for rectification.

\section{Additional Dataset Description}

\subsection{Data Capturing Steup}
We captured our dataset with a volumetric capture system~\cite{collet2015high} equipped with 106 synchronized cameras (53 RGB and 53 IR cameras). The sequences are filmed at 12 MP, 30 FPS, and within an effective capture volume of 2.8 m in diameter and 3 m in height. Each frame consists of a mesh with 80k faces and a texture map.

\begin{figure}[t]
\centering
\includegraphics[width=1\linewidth]{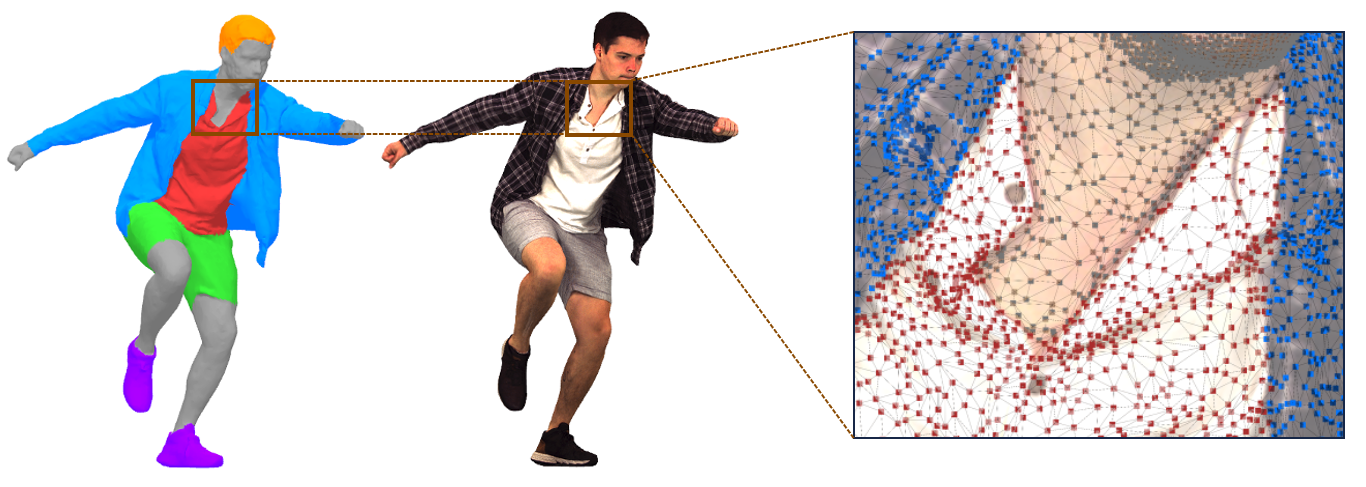}
\vspace{-1em}
\caption{\textbf{Vertex-level semantic annotations.} Our dataset contained precise vertex-level semantic labels of clothing categories. }
\label{fig:supp_quality}
  \vspace{-1.em}
\end{figure}

\subsection{Dataset Contents}

Our 4D-DRESS dataset provides the following data, examples are shown in \cref{fig:supp_dataset}:
\begin{itemize}
    \item \textbf{4D textures scans.} High-quality 4D textured scans of 32 subjects, 64 human outfits (32 Inner and 32 Outer), with 520 motion sequences and 78k frames in total.
    \item \textbf{Vertex-level annotations.} We offer accurate vertex-level annotations through our 4D human parsing pipeline. An example of our label quality is shown in \cref{fig:supp_quality}. Using these labels, we also provide multi-view images with semantic labels in 2D.
    \item \textbf{Parametric body models.} We register precise SMPL and SMPL-X body models for each frame.
    \item \textbf{Garment meshes}. We extract 3D garment meshes based on the vertex labels.
\end{itemize}

\begin{figure}[t]
\centering
\includegraphics[width=\linewidth]{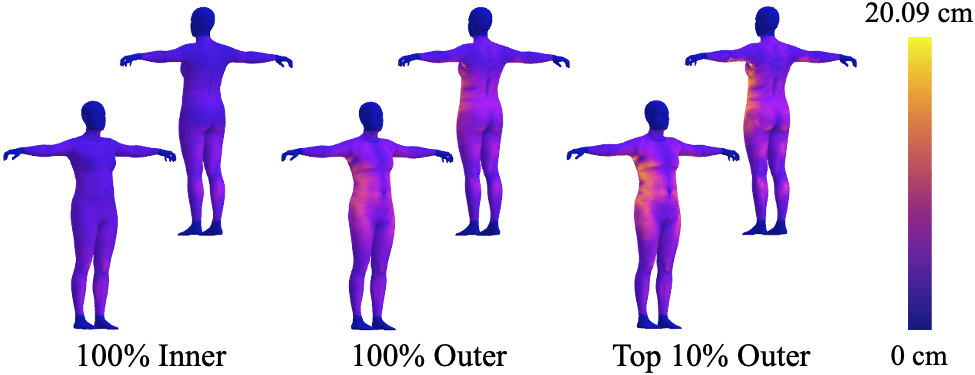}
\caption{\textbf{Visualization of 4D-DRESS outfits distance}. The mean distance distribution from garment outfits to SMPL bodies.}
  \vspace{-1.5em}

\label{fig:dataset_outfit}
\end{figure}

\begin{figure}[t]
\centering
\includegraphics[width=\linewidth]{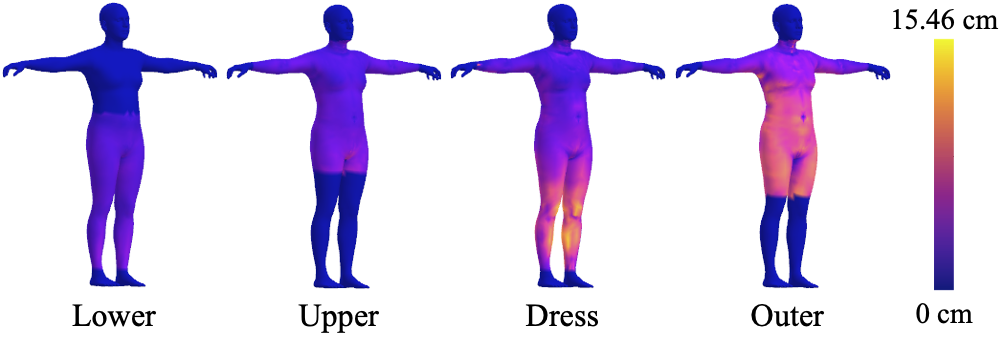}
\caption{\textbf{Visualization of 4D-DRESS outfits distance}. The mean distance distribution from garment meshes to SMPL bodies.}
  \vspace{-1.5em}

\label{fig:dataset_cloth}
\end{figure}
\subsection{Clothing Distribution}
We compute the mean distances from the outfits to the registered SMPL body surfaces. The inner and outer outfits exhibit distance ranges of up to 7.12 cm and 14.76 cm, respectively, over all frames. The distribution of the distance on the SMPL body is shown in \cref{fig:dataset_outfit}. In the 10\% most challenging frames that have a larger Chamfer distance between scan mesh and SMPL mesh, the distance range increases to 20.09 cm for outer outfits. We further visualize the mean distances of each garment category, as shown in \cref{fig:dataset_cloth}. The average Chamfer distance between the clothed human scans and SMPL body meshes are 3.30 cm and 5.28 cm for the inner and outer outfits in our {\datasetname} dataset, and 2.21 cm in the X-Humans dataset \cite{X-Avatar}.
\section{Experimental Details}

\subsection{Clothing Simulation}
\label{sec:bench:cloth_simulatnon}
4D-Dress provides diverse garments and challenging human pose sequences, which serves as a great asset for future research in clothing simulation. Unlike the synthesized garment templates with smooth surfaces and simple topologies, we provide templates extracted from scans, with realistic wrinkles and complex structures. Using these templates, we evaluated the performance of recent unsupervised cloth simulators, including PBNS~\cite{PBNS}, Neural Cloth Simulator (NCS)~\cite{NCS} and HOOD~\cite{HOOD}, and a baseline method, linear blend-skinning. We quantitatively and qualitatively compared the generated garments with our scanned garments. We also demonstrated the potential of HOOD by simply optimizing the material parameters, which again confirmed the value of our dataset.
In the following sections, we elaborate on each step of our experiments.

\subsection{Template Extraction}
\label{sec:bench:template_extract}
Current clothing simulation algorithms rely on a predefined garment template, deforming it to generate realistic simulations under various poses. They typically utilized synthesized garment templates, with unnaturally smooth surfaces and basic topologies. In our work, we provided templates directly extracted from real-world scans, offering a more realistic foundation for deformation.

Firstly, we select from pose sequences the frames closest to the canonical pose, in other words, “T-pose”. We also make sure that the body in this frame is static and garments are in rest status. Then we apply inverse LBS to convert the scans into exact canonical pose. After extracting garment meshes from the unposed scans, we made some manual efforts to recover the garment shape in Blender~\cite{Blender}. Specifically, we erased unwanted faces, solved penetrations between clothing and body, and smoothed rigid wrinkles and coarse boundaries. Synthesized templates used by current simulators usually have 4-5k vertices. We observed in experiments that too many vertices in the template are computationally expensive for simulation and may erode performance. Therefore, we downsampled each template to 30-50\%, which now has 3-8k vertices in total depending on each garment's surface area, while keeping them in their original shapes. To use lower garments in simulators, like pants and lower skirts, pinned vertices are compulsory for them to stay on the body. We extract the loop around the waist as pinned vertices and provide their indexes. 

\begin{figure}[t]
\centering
\includegraphics[width=1.\linewidth]{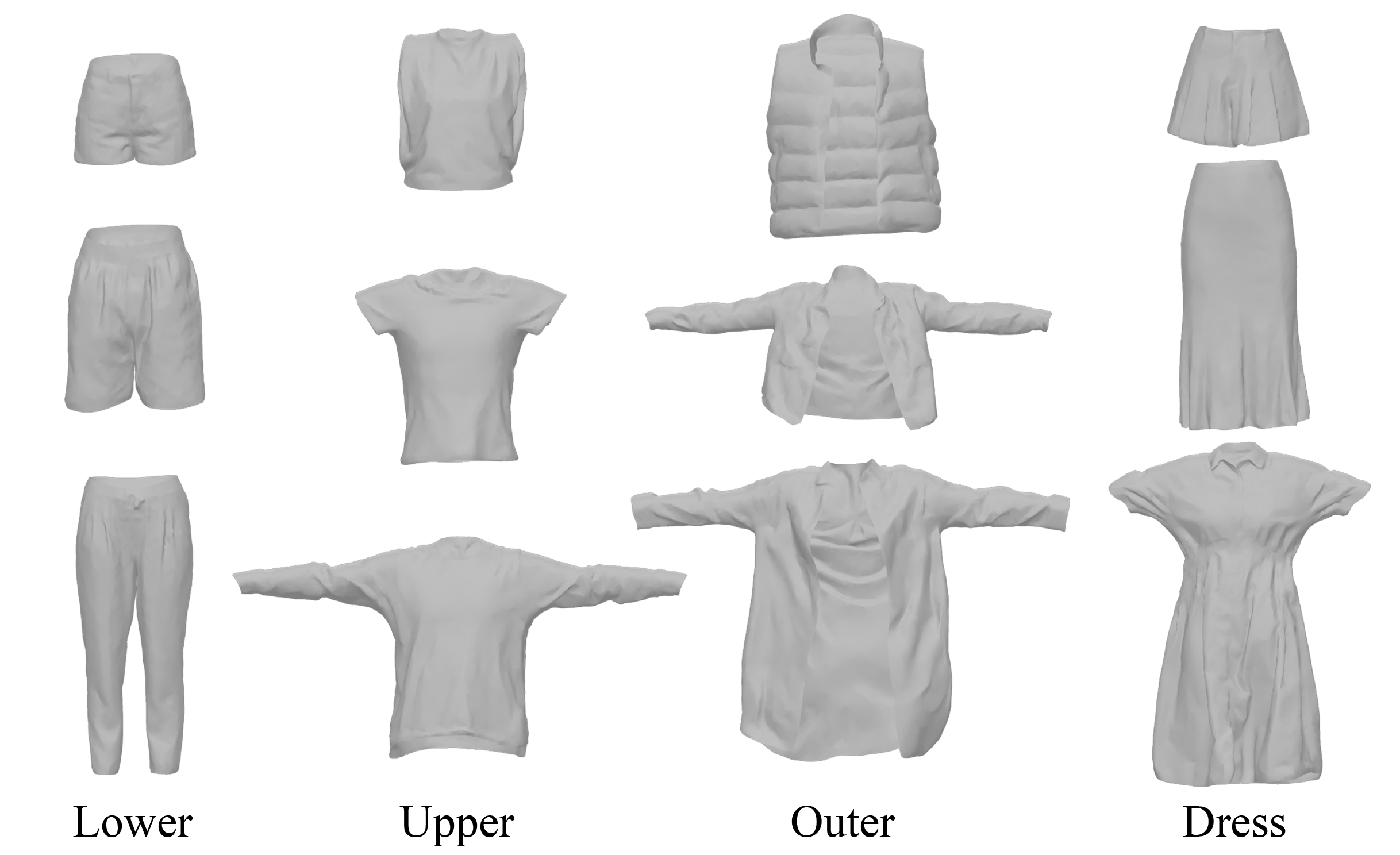}
\vspace{-1em}
\caption{\textbf{Garment templates used for clothing simulation.} We extract four types of Garment templates from T-pose scans.}
\label{fig:supp_templates}
  \vspace{-1.em}
\end{figure}

\subsection{Evaluation Details}
\label{sec:bench:simulator_evaluation}
In the clothing simulation benchmark, we compared four different clothing simulators: LBS, PBNS~\cite{PBNS}, NCS~\cite{NCS}, and HOOD~\cite{HOOD}. The training and evaluation of each method were conducted using the SMPLX model, which provides more details in visualization. The final evaluation is done on four types of garments(Upper, Outer, Dress, and Lower), with each having 2 garments and 6 sequences in total. For qualitative evaluation, we employed Chamfer distance and stretching energy, scaling vertex positions by a factor of 100 to use centimeters as the unit.

The Chamfer distance, shown in equation \ref{eq:eval_chamfer}, is computed by summing the squared distances between nearest neighbor correspondences of two-point clouds. We denote the sampled points on simulation and ground-truth meshes as $X$ and $Y$, respectively, with $N_*$ representing the amount of sampled points, set to 100,000 in our experiment.

\begin{equation}\label{eq:eval_chamfer} 
d_{CD} = \frac{1}{N_x}\sum_{x\in X} \underset{y\in Y}{min}\| x - y\|_2^2 +  \frac{1}{N_y}\sum_{y\in Y} \underset{x\in X}{min}\| x - y\|_2^2
\end{equation}

\begin{figure*}[t]
\centering
\includegraphics[width=1\linewidth]{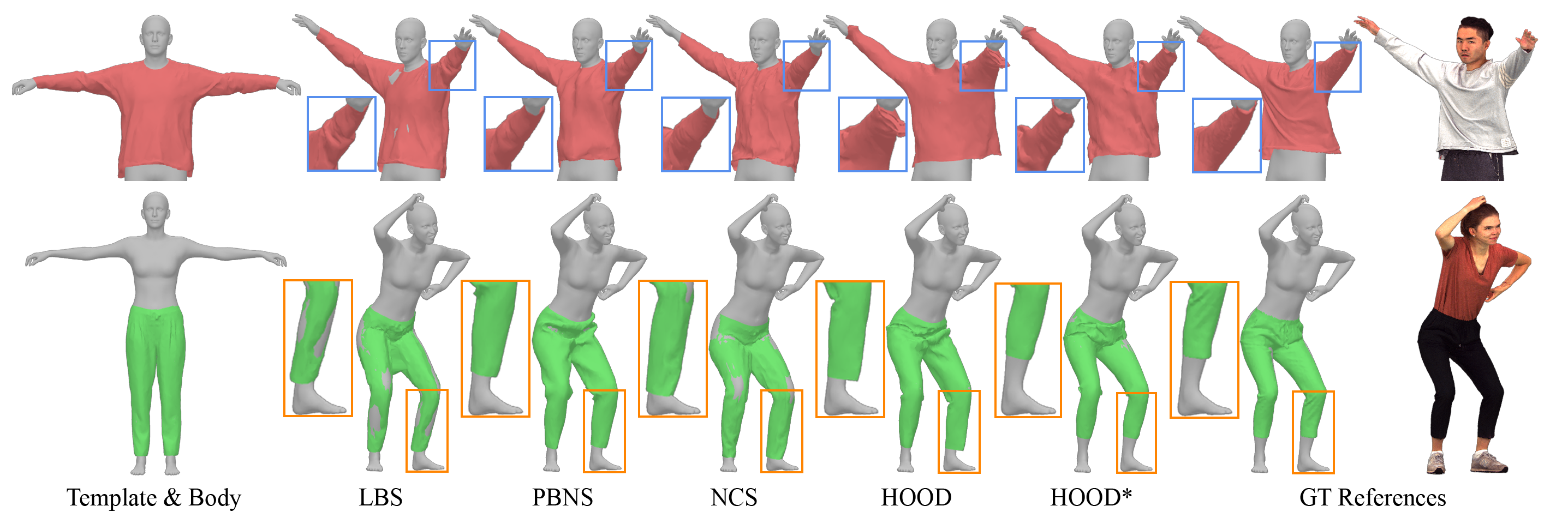}
\vspace{-1em}
\caption{\textbf{Additional qualitative results for clothing simulation.} Left are templates used for simulations. Right are simulations and ground-truth scans. HOOD presents more dynamic while getting overly stretched. HOOD* matches well with ground truth.}
\label{fig:supp_simulation}
  \vspace{-1.em}
\end{figure*}

The stretching energy, widely used in mass-spring-based simulators, is computed as equation \ref{eq:eval_stretching}, where $N_e$ is the total number of edges, ${e}_{i}$ and $ \bar{e}_{i}$ are the lengths of the edge $i$ in the current frame and the template respectively. 

\begin{equation}\label{eq:eval_stretching}
E_{str} = \frac{1}{N_e}\sum_{i} \|{e}_{i} - \bar{e}_{i}\|^2
\end{equation}

We provide more details on implementing each method: 
\textbf{LBS} blends joint transforms with skinning-weights. For each garment template, we find the nearest body node on the canonical SMPLX human, and get the skinning weights on this point. Then, we follow the same forward LBS process in SMPLX to get deformed template meshes.

\textbf{PBNS} and \textbf{NCS}, both are deformation-based methods, predict vertex-wise deformation on the template and employ LBS to transform the deformed garment into desired poses. Given their "One model for one garment" nature, we trained each garment from scratch. We also used identical AMASS sequences mentioned in the NCS paper to ensure fairness. As both PBNS and NCS developed using SMPL, we made slight adjustments to the data-loading pipeline to ensure their compatibility with SMPLX. And we assigned zero poses to joints that are exclusive in SMPLX.

Meanwhile, we also kept the same training settings used in their original papers. For PBNS, default parameters were used, and each garment underwent training for 20-50 epochs to ensure convergence. For NCS, a batch size of 2048 was employed across all training instances, as suggested in their paper. In the case of tight garments, default parameters were maintained with a temporal window size of 0.5 and 10 iterations for blend weights smoothing. In the case of loose garments like outerwear and dresses, we made slight parameter adjustments for stable training, typically using a temporal window size of 0.75 and 1, with 50 iterations for blend weights smoothing, as suggested by the author in a GitHub issue.

\textbf{HOOD}, as a simulation-based method, predicts physically realistic fabric dynamics and is agnostic to garment topology. Hence, we directly used a pre-trained publicly available model to evaluate our garments. Unlike the deformation-based methods, which convert the template in canonical pose to any pose instantly, HOOD predicts garment motion frame by frame. Therefore, to apply our canonical template for simulating each sequence, we have to convert the template into the pose of the first frame. In the HOOD paper, they used LBS to convert templates, which works for tight synthesized garments. However, for our real-world garments, it usually results in large stretching on mesh, especially around joint areas. Therefore, alternatively, we insert extra frames from the canonical pose to the first frame and simulate the prolonged sequence to get a natural transform from the canonical pose. The first poses for all sequences in our dataset are in A-pose. Generally, we insert 30 frames to transfer from canonical to A-pose, which makes it slow enough for the garment to stay in rest status with minimum dynamics.

\subsection{HOOD*: Material Optimization}
\label{sec:bench:HOOD*}
HOOD provides 4 local material parameters for each vertex, including $\mu$ and $\lambda$ evaluating the ability of stretching and area preservation, mass $m$ computed from the fabric density, and the bending coefficient $k_{bending}$ penalizing folding and wrinkles. For each edge, there are three material parameters, including  $\mu$, $\lambda$, and $k_{bending}$. Assuming we have $v$ vertices, $e$ edges, and coarse edges in total, we define the material parameters as $\mathcal{M}\in\mathbb{R}^{4v+3e}$.

In the fine-tuning process, we freeze the pre-trained HOOD model $\mathcal{H}$ and only update material parameters $\mathcal{M}$. Using all 6 sequences of each garment for training, we feed them into model $f$ to get simulated outputs. Then, with Ground Truth garment mesh $G$, we compute Chamfer distance and stretching energy, as described in equation \ref{eq:material_finetuning}. 

\begin{equation}\label{eq:material_finetuning}
    \mathcal{L} = \mathcal{L}_{CD}(f(\mathcal{M},\mathcal{H}), G) + w \mathcal{L}_{Estr}(f(\mathcal{M},\mathcal{H}), G)
\end{equation}

We used the stretching energy from HOOD and set $w$ as 1 in our experiments. Chamfer distance $\mathcal{L}_{CD}$ is described in equation \ref{eq:chamfer_dist}, measuring the average distance between simulation and ground-truth garment. We use $V_*,\ (*\in[s,g])$ to represent the simulated and ground truth vertices and use $N_*$ as the total number of vertices.

\begin{equation}\label{eq:chamfer_dist}
\mathcal{L}_{CD} = \frac{1}{N_s}\sum_{x\in V_s} \underset{y\in V_g}{min}\| x - y \|^2 +  \frac{1}{N_g}\sum_{y\in V_g} \underset{x\in V_s}{min}\| x - y \|^2
\end{equation}

For each garment, we trained with Adam Optimizer with a learning rate of 5e-4. And it usually takes 50 epochs to converge. Generally, HOOD* gets a much lower distance compared to ground truth mesh quantitatively, and also performs more natural fabric dynamics qualitatively.

\label{sec:bench:simultion_details}

\end{document}